%% file: main.tex
\documentclass[11pt]{article}
\usepackage{autobci_arxiv}
\usepackage[T1]{fontenc}
\usepackage[utf8]{inputenc}
\usepackage{amsmath,amssymb,graphicx,booktabs,microtype}
\usepackage{xcolor}
\DeclareRobustCommand{\rejectionrisk}[1]{#1}

\usepackage{colortbl,tabularx}
\usepackage{wrapfig}
\usepackage{float}
\usepackage{flafter}
\newlength{\tablebottomtrim}
\AddToHook{env/table/end}{\par\vspace{-\tablebottomtrim}}
\AddToHook{env/table*/end}{\par\vspace{-\tablebottomtrim}}
\usepackage{listings}
\usepackage{etoc}
\usepackage[normalem]{ulem}
\lstdefinestyle{prompt}{basicstyle=\ttfamily\scriptsize,
  breaklines=true,breakatwhitespace=false,columns=fullflexible,
  keepspaces=true,showstringspaces=false,frame=single,
  rulecolor=\color{black!20},backgroundcolor=\color{black!2},
  aboveskip=0.8em,belowskip=0.8em}
\usepackage{hyperref}
\hypersetup{colorlinks=true,linkcolor=preprintblue,citecolor=preprintblue,urlcolor=preprintblue,pdftitle={AutoBCI}}

\title{AutoBCI: Forecast-Guided Agentic Neural Architecture Discovery for EEG-Based Brain-Computer Interfaces}
\author{\normalfont\small Muyun Jiang$^{1}$, Yi Ding$^{1}$, Wei Zhang$^{1}$, Jinbo Chen$^{1}$, Chenyu Liu$^{1}$, Zhenjie Yang$^{2}$,\\
\normalfont\small Yuxin Li$^{1}$, Jingyuan Chen$^{1}$, Yuhao Lu$^{1}$, Yong Li$^{3}$, Shuailei Zhang$^{1}$, Cuntai Guan$^{1}$\\[0.5em]
\normalfont\footnotesize\color{preprintgray} $^{1}$Nanyang Technological University, Singapore\\
\normalfont\footnotesize\color{preprintgray} $^{2}$The University of Hong Kong, Hong Kong, China\\
\normalfont\footnotesize\color{preprintgray} $^{3}$Southeast University, China}
\hypersetup{pdfauthor={Muyun Jiang, Yi Ding, Wei Zhang, Jinbo Chen, Chenyu Liu, Zhenjie Yang, Yuxin Li, Jingyuan Chen, Yuhao Lu, Yong Li, Shuailei Zhang, Cuntai Guan}}

\begin{document}
\raggedbottom
\maketitle

\begin{center}
  \includegraphics[width=\linewidth]{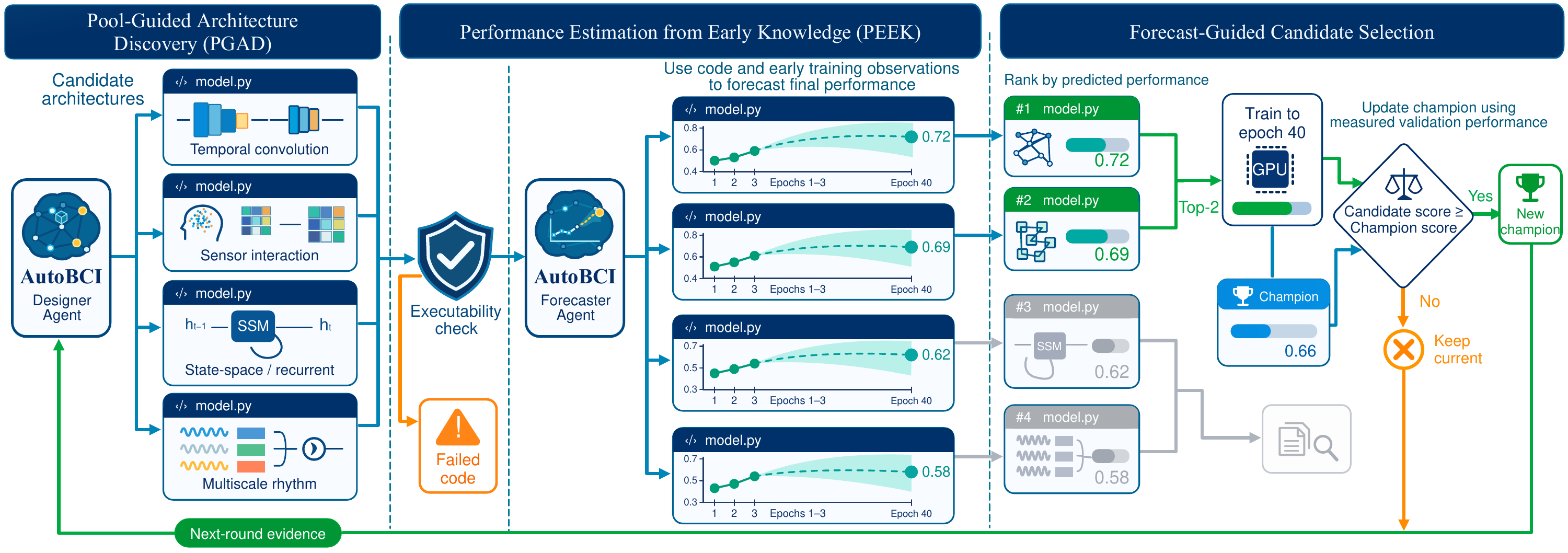}
  \captionof{figure}{\textbf{AutoBCI framework overview.}  The Designer Agent uses PGAD to generate and refine EEG architectures;  the Forecaster Agent uses PEEK to predict performance and guide  continued training. Validation feedback guides subsequent discovery  rounds.}
  \label{fig:overview}
\end{center}

\begin{abstract}
EEG-based brain--computer interfaces support a broad range of applications, yet designing decoding architectures that perform well across diverse tasks remains challenging. We introduce AutoBCI, an agentic framework in which a Designer Agent and a Forecaster Agent support the discovery and selection of EEG decoding architectures across tasks. The Designer Agent performs Pool-Guided Architecture Discovery (PGAD), generating and refining architectures through training and validation across multiple EEG tasks, such as emotion recognition, motor imagery, and sleep staging. \textcolor{black}{The Forecaster Agent performs Performance Estimation from Early Knowledge (PEEK), using architecture code, the training protocol, and early learning curves to predict full-budget validation performance and select promising candidates for continued training.} Across 14 EEG datasets spanning motor imagery, emotion recognition, and sleep staging, we evaluate AutoBCI with six LLMs, including Opus 5.5 and GPT 5.6 Sol, and compare the architectures selected by the search procedure against ten baselines: six conventional EEG models and four foundation models.
The architecture discovered by AutoBCI with Claude Opus 5.5 achieves 64.16\% average test balanced accuracy (bAcc), compared with 63.87\% for REVE, the strongest baseline on this metric. \textcolor{black}{Using ten observed epochs, PEEK reduces mean absolute error in predicting average validation bAcc from 2.20 to 1.36 percentage points, a 38.1\% reduction relative to the best-observed-score baseline.}
Code is available at \url{https://github.com/Jiang-Muyun/AutoBCI}.
\end{abstract}

\section{Introduction}
\label{sec:introduction}

Electroencephalography (EEG) is a key modality for non-invasive brain--computer interfaces. EEG decoding spans tasks with different spatial coverage, temporal scales, and label spaces. For over a decade, researchers have encoded domain knowledge through selected frequency bands, temporal windows, and spatial filters, such as common spatial patterns for motor-imagery decoding \citep{schirrmeister2017deep}. Deep learning learns features directly, but model structures still reflect human design: EEGNet uses compact spatiotemporal convolutions \citep{lawhern2018eegnet}, TSception captures multiple temporal scales and spatial asymmetry \citep{ding2022tsception}, and DeepSleepNet models sleep-stage dependencies with convolutional and recurrent layers \citep{supratak2017deepsleepnet}. These advances demonstrate \textbf{the value of EEG-specific architectural design}, while \textbf{selecting suitable structures across tasks remains a manual burden}. This motivates automated discovery of architectures that perform well across different EEG tasks.

Agentic AI offers a way to reduce the manual effort involved in working across these settings. Existing systems demonstrate that \textbf{agents can automate EEG analysis workflows}. EEGAgent supports signal exploration, event detection, and report generation \citep{zhao2026eeg}; CogEEGAgent translates user questions into registered analyses with independent confirmation \citep{hou2026cogeegagent}; and NS-Copilot coordinates specialized agents and pretrained models for neuroscience workflows \citep{liu2026ns}. These capabilities improve access to analysis tools, but \textbf{\uline{these systems focus on automating analysis rather than designing decoding architectures}}. Reducing the model-design burden requires extending this automation to the architectures themselves, using experimental feedback to guide structural changes.

Work on neural architecture search takes this further, showing that \textbf{architecture design can be automated through search and feedback}. CTNAS-EEG introduces a compatible search space and constrained search procedure across EEG tasks \citep{duan2023cross}. NeuroWeaver evolves executable EEG pipelines using domain-informed initialization and performance, novelty, and efficiency objectives \citep{wang2026neuroweaver}. More broadly, EvoPrompting evolves architecture code, while NADER uses collaborating agents to guide architectural modifications \citep{chen2023evoprompting,yang2025nader}. However, \textbf{\uline{success on separately optimized tasks does not establish the quality of a shared architecture}}. Our goal is to discover a single architecture that performs well across EEG tasks. We therefore train and validate each candidate separately for each task, then use the combined validation results to guide architecture selection and refinement. In this study, we evaluate this approach on three common EEG tasks: emotion recognition, motor imagery, and sleep staging. Evaluating each proposal across these tasks increases the training cost of discovery, making candidate screening an important part of the search.

To reduce this evaluation cost, prior work exploits the finding that \textbf{partial training evidence can support early performance prediction}. Learning-curve extrapolation identifies unpromising runs \citep{domhan2015speeding}, and performance predictors combine architecture features, hyperparameters, and partial validation trajectories \citep{baker2017accelerating}. LLM-ENAS-PKB further uses language models as architecture-performance surrogates \citep{weilin2026large}. However, \textbf{\uline{forecast accuracy alone does not establish reliable candidate selection}}. For screening to be useful, promising candidates must remain in the search. We therefore use architecture code and early learning curves to predict how well a candidate will perform after full training. These forecasts help identify promising candidates for continued training. We evaluate both prediction accuracy and whether the selected candidates retain the architectures that perform best after full training.

To address these challenges, we introduce \textbf{\uline{AutoBCI}}, an agentic framework containing two agents: a Designer Agent that uses Pool-Guided Architecture Discovery (PGAD) to generate and refine EEG architectures, and a Forecaster Agent that uses Performance Estimation from Early Knowledge (PEEK) to predict candidate performance and guide continued training, as shown in Figure~\ref{fig:overview}. First, \textbf{PGAD automates architecture design through executable code generation and refinement}. The agent proposes models and uses measured validation feedback to revise their structure, extending its role from analysis workflow execution to model design. Second, \textbf{PGAD selects shared architectures using joint evidence across tasks}. Each candidate is trained and validated independently on emotion recognition, motor imagery, and sleep staging. An equal-task validation objective ranks a persistent candidate pool, from which strong parents are refined alongside fresh proposals. Third, \textbf{PEEK guides training allocation using early performance forecasts}. It combines architecture code, the training protocol, and early learning curves to predict full-budget validation scores and rank candidates for continued training. These mechanisms connect model generation, cross-task evaluation, and candidate screening within one discovery framework.

Our contributions are threefold. First, we introduce AutoBCI and develop PGAD for agentic EEG architecture discovery, using a persistent candidate pool and cross-task validation feedback to guide code generation and refinement. Second, we develop PEEK to forecast full-budget performance from architecture code and early learning curves, enabling candidate screening before training is complete. Third, we evaluate six LLM-guided searches across 14 datasets and compare their champions with ten baselines. The searches produce 269 architectures trained on all three task families; the strongest champion achieves 64.16\% average test bAcc, exceeding the strongest baseline by 0.29 percentage points. PEEK reduces forecast mean absolute error by 38.1\% after ten epochs. Retaining three candidates per round preserves a full-budget winner in 91.7\% of rounds, with an estimated 44.9\% reduction in training epochs.

\section{Method}
\label{sec:method}

AutoBCI combines a Designer Agent for architecture discovery with a Forecaster Agent for early candidate selection. The Designer Agent uses PGAD to generate architectures, evaluate separately trained instances across EEG tasks, and refine candidates using their combined validation scores. The Forecaster Agent uses PEEK to predict full-budget validation performance from architecture code, the training protocol, and early learning curves, helping select candidates for continued training.

\subsection{Pool-Guided Architecture Discovery}
\label{sec:had}

PGAD organizes architecture discovery around a persistent pool of evaluated candidates. Starting from an initial set of proposals, the Designer Agent uses validation feedback across tasks to identify strong architectures and propose refinements. Each subsequent round combines these refinements with fresh designs, allowing the search to build on previous discoveries while exploring alternatives. The procedure consists of candidate generation, cross-task evaluation, and pool-based selection and refinement, as detailed below.

\paragraph{Candidate generation.}
PGAD runs for $R$ discovery rounds, indexed by $r\in\{1,\ldots,R\}$, with $N$ candidate architectures proposed per round. Round $r=1$ contains $N$ fresh proposals from the configured LLM. Each proposal contains complete PyTorch model code and a short architectural hypothesis. The model interface accepts configurable channel, sample, and class counts, mapping a batch of EEG signals to class logits. Before training, the framework checks the response format and model interface. Candidate code is preserved as generated; implementation errors are recorded as candidate failures.

\paragraph{Evaluation across tasks.}
For each candidate, the same architecture source is trained independently on $K$ task families, indexed by $k\in\{1,\ldots,K\}$, with separate model parameters and checkpoints for each task. Our evaluation uses emotion recognition, motor imagery, and sleep staging ($K=3$). Let $\mathcal{D}_k$ denote the datasets in task family $k$, $E$ the full training budget in epochs, and $b_{a,d}(e)$ denote the validation balanced accuracy of architecture $a$ on dataset $d$ at epoch $e$. At each epoch, datasets within a task receive equal weight. The task score is the highest such mean over the training run:
\begin{equation}
  s_{a,k}=\max_{1\leq e\leq E}\frac{1}{|\mathcal{D}_k|}
    \sum_{d\in\mathcal{D}_k} b_{a,d}(e).
  \label{eq:task-score}
\end{equation}
Thus, all datasets within a task share one selected checkpoint epoch, while different tasks may select different epochs. The architecture score assigns equal weight to the $K$ task scores:
\begin{equation}
  S_a=\frac{1}{K}\sum_{k=1}^{K}s_{a,k}.
  \label{eq:architecture-score}
\end{equation}
An architecture is eligible for selection only after its training runs on all $K$ tasks complete $E$ epochs.

\begin{figure}[!t]
  \centering
  \begin{minipage}[t]{0.60\linewidth}
  \vspace{0pt}
  \includegraphics[width=\linewidth]{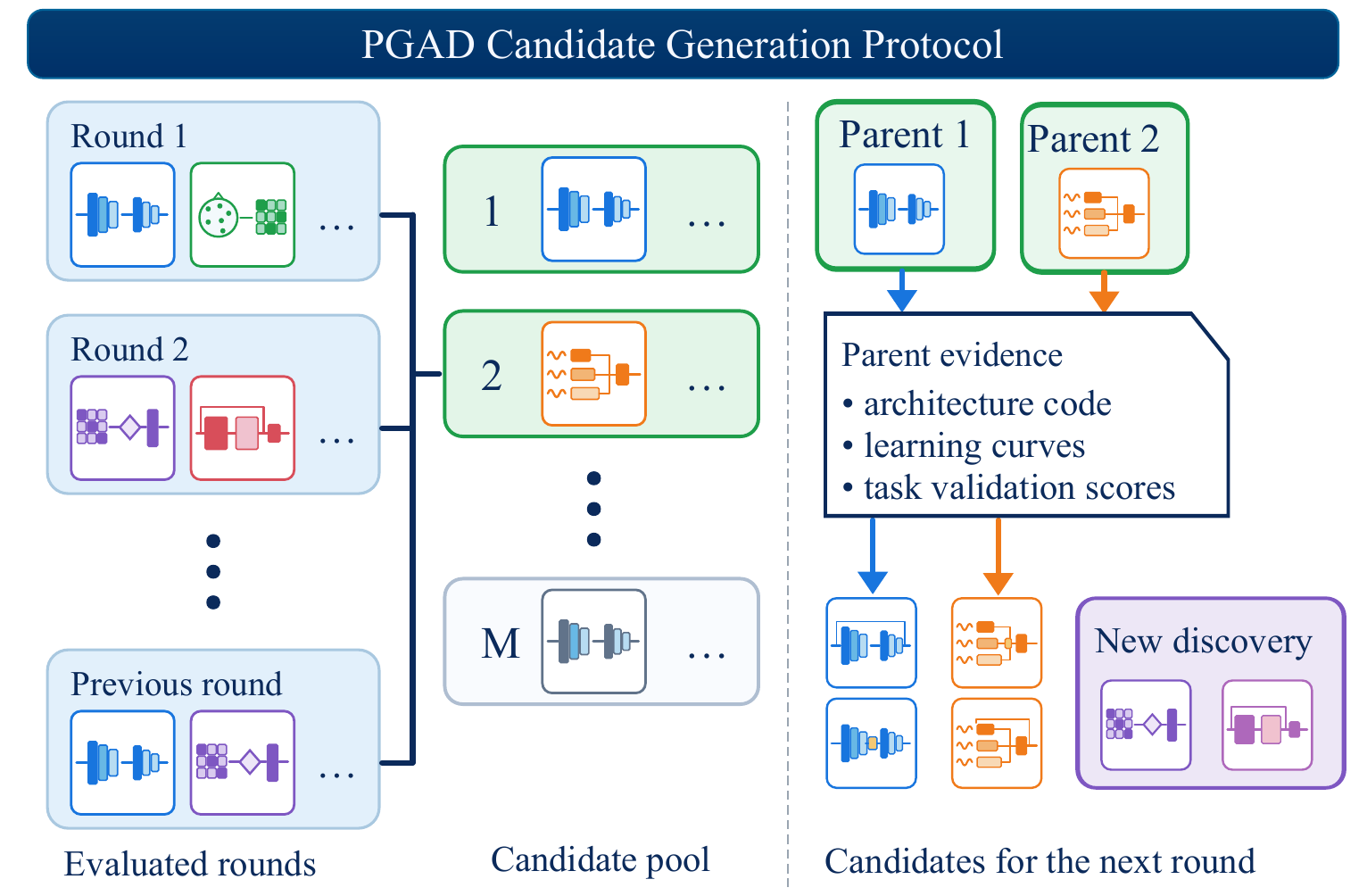}
  \end{minipage}\hfill
  \begin{minipage}[t]{0.37\linewidth}
  \vspace{0pt}
  \caption{\textbf{PGAD candidate generation protocol.}  \textbf{Left:} Architectures evaluated in previous discovery rounds.  \textbf{Middle:} A shared candidate pool ranked by combined validation  performance. \textbf{Right:} Top-ranked parents provide source code,  learning curves, and task-specific validation scores for targeted  refinements. These refinements are combined with fresh architecture  proposals to form the next candidate batch.}
  \label{fig:pool}
  \end{minipage}
\end{figure}

\paragraph{Selection and refinement.}
Before round $r>1$, eligible architectures from all preceding rounds form a shared pool $\mathcal{P}_{r-1}$ of size $M_{r-1}=|\mathcal{P}_{r-1}|\leq N(r-1)$, as illustrated in Figure~\ref{fig:pool}. The round selects the $P$ highest-scoring architectures by $S_a$ as parents and requests $Q$ refinements per parent. Each refinement receives the parent's source code, learning curves, and task-specific validation results, and is prompted to change one architectural component and describe the intended change. Alongside these $PQ$ refinements, the LLM generates $U$ fresh explorations, with
\begin{equation}
  N=PQ+U.
  \label{eq:proposal-budget}
\end{equation}
Thus, the search requests $RN$ architectures over $R$ rounds. The highest-scoring eligible architecture in $\mathcal{P}_R$ is the final champion. Validation data determine parent and champion selection; the held-out test split is used only for a separate final evaluation.

\subsection{Performance Estimation from Early Knowledge (PEEK)}
\label{sec:peek}

PEEK uses early performance forecasts to select promising architectures for continued training. After a short initial training period, the Forecaster Agent combines each candidate's architecture code, training protocol, and observed learning curves to estimate its full-budget validation performance. Architecture code describes the model's design, while the curves show how that design learns on the target task. 
PEEK uses this complementary evidence to rank candidates and allocate the remaining training budget to the predicted leaders. The procedure has two steps: performance forecasting and candidate selection.

\paragraph{Performance forecasting.}
For task $k$, each candidate $a$ first trains for $e_0<E$ epochs. Let $C_a$ denote its source code and $\mathcal{H}_{a,k}^{1:e_0}$ its observed training and validation history. Given the training protocol $\pi_k$, PEEK produces
\begin{equation}
  \widehat{s}_{a,k}
  =F_{\mathrm{LLM}}\!\left(C_a,\mathcal{H}_{a,k}^{1:e_0},\pi_k\right),
  \label{eq:peek-forecast}
\end{equation}
where the target $s_{a,k}$ is the best equal-dataset validation balanced accuracy at a common checkpoint within $E$ epochs (Equation~\eqref{eq:task-score}). Aligning the forecast with this selection criterion lets PEEK assess a candidate's potential over the remaining budget, including improvement beyond its early score. Predictions lie in $[0,1]$ and must be at least the best validation score already observed. The LLM receives only candidate code, the specified protocol, and the observed portion of the learning history. For shared-architecture selection, we average the task forecasts using the same equal-task weighting as Equation~\eqref{eq:architecture-score}:
\begin{equation}
  \widehat{S}_a=\frac{1}{K}\sum_{k=1}^{K}\widehat{s}_{a,k}.
  \label{eq:combined-forecast}
\end{equation}

\paragraph{Candidate selection.}
To select shared architectures, PEEK ranks candidates by the combined forecast $\widehat{S}_a$ (Equation~\eqref{eq:combined-forecast}). The screening policy retains the $N_{\mathrm{keep}}$ highest-ranked candidates and continues each on all $K$ tasks to $E$ epochs, where $1\leq N_{\mathrm{keep}}\leq N$. Only candidates that complete all $K$ tasks enter the PGAD pool, using their measured scores $S_a$; partially trained candidates remain ineligible for parent and champion selection. For $N$ candidates initially trained for $e_0$ epochs on each task, the nominal cost is $K[Ne_0+N_{\mathrm{keep}}(E-e_0)]$ task-epochs, compared with $KNE$ for full training. The retention--cost analysis sweeps $N_{\mathrm{keep}}=1,\ldots,6$, with the abstract reporting the three-candidate operating point. Separate task-level diagnostics rank candidates by $\widehat{s}_{a,k}$ and measure top-one and top-two retention for each task.

\begin{table}[!t]
  \centering
  \caption{EEG datasets, task-specific inputs, and class counts.}
  \label{tab:dataset-splits}
  \fontsize{8}{9.5}\selectfont
  \setlength{\tabcolsep}{3pt}
  \renewcommand{\arraystretch}{1.0}
  \setlength{\aboverulesep}{0.2ex}
  \setlength{\belowrulesep}{0.2ex}
  \def\datasetno#1{\hspace*{.5\linewidth}\makebox[0pt][c]{#1}}
  \begin{tabularx}{\linewidth}{@{}>{\centering\arraybackslash}p{0.11\linewidth}p{0.05\linewidth}>{\raggedright\arraybackslash}Xp{0.13\linewidth}p{0.28\linewidth}@{}}
    \toprule
    Task & \datasetno{No.} & Datasets & Unified input & Classes \\
    \midrule
    \raisebox{-4\baselineskip}[0pt][0pt]{\shortstack[c]{Motor\\Imagery}} &
    \datasetno{1}\newline \datasetno{2}\newline
    \datasetno{3}\newline \datasetno{4}\newline
    \datasetno{5}\newline \datasetno{6}\newline
    \datasetno{7}\newline \datasetno{8} &
    BCIC-IV-2a \citep{tangermann2012review}\newline
    OpenBMI-MI \citep{lee2019eeg}\newline
    BCIC-Upperlimb \citep{jeong20222020}\newline
    Cho2017 \citep{cho2017eeg}\newline
    HighGamma \citep{schirrmeister2017deep}\newline
    PhysioNet-MI \citep{schalk2004bci2000}\newline
    SHU-MI \citep{ma2022large}\newline
    Shin2017A \citep{shin2016open} & \raisebox{-3.5\baselineskip}[0pt][0pt]{65 ch $\times$ 4 s} & \raisebox{-3\baselineskip}[0pt][0pt]{\parbox[t]{\linewidth}{Left, Right, Foot, Tongue, Cylin, Sphe, Lumbrical (7 classes)}} \\
    \midrule[0.3pt]
    \raisebox{-1.5\baselineskip}[0pt][0pt]{\shortstack[c]{Emotion\\Recognition}} &
    \datasetno{9}\newline \datasetno{10}\newline \datasetno{11} &
    SEED \citep{duan2013differential}\newline
    SEED-IV \citep{zheng2018emotionmeter}\newline
    SEED-V \citep{liu2021comparing} & \raisebox{-\baselineskip}[0pt][0pt]{65 ch $\times$ 4 s} & \raisebox{-.5\baselineskip}[0pt][0pt]{\parbox[t]{\linewidth}{Positive, Neutral, Negative, Sad, Fear, Happy, Disgust (7 classes)}} \\
    \midrule[0.3pt]
    \raisebox{-1.5\baselineskip}[0pt][0pt]{\shortstack[c]{Sleep\\Staging}} &
    \datasetno{12}\newline \datasetno{13}\newline \datasetno{14} &
    Sleep-EDF \citep{kemp2000analysis}\newline
    ISRUC \citep{khalighi2016isruc}\newline
    HMC \citep{alvarez2021inter}
    & \raisebox{-\baselineskip}[0pt][0pt]{8 ch $\times$ 30 s} & \raisebox{-\baselineskip}[0pt][0pt]{Wake, N1, N2, N3, REM (5 classes)} \\
    \bottomrule
  \end{tabularx}
\end{table}

\section{Experimental Setup}
\label{sec:setup}

\textcolor{black}{Our experiments evaluate architecture discovery and early candidate selection on shared EEG datasets. PGAD evaluates candidates with a 40-epoch budget. Completed runs establish the full-budget scores and winners needed to evaluate PEEK. The forecaster receives only architecture code, the training protocol, and early learning curves; full-run outcomes provide ground truth for retrospective evaluation of forecast accuracy and winner retention.}

\subsection{Dataset construction}
\label{sec:datasets}

EEG tasks differ in their prediction targets and recording characteristics, so we organize the 14 datasets into motor imagery (MI), emotion recognition (ER), and sleep staging (SS), as shown in Table~\ref{tab:dataset-splits}. Datasets within each family are pooled to train one model instance per task. Channel layouts and labels are harmonized within a task. MI and ER recordings are mapped to the 65-channel standard 10--10 system template used by LEAF \citep{jiang2025leaf}, while the sparser sleep recordings use an eight-channel template. Validation and test scores are reported separately for each dataset.

All candidates use the same fixed training, validation, and test partitions and sleep subsets. Training fits model parameters, validation selects checkpoints and architectures, and held-out test data are used only for final evaluation. Preprocessing and split construction are detailed in Supplementary~\ref{sec:data-preparation}.

\subsection{Architecture search protocol}
\label{sec:search-setup}

On these datasets, PGAD searches share a common budget to support comparison of architecture discovery across LLMs. Each search comprises $R=6$ discovery rounds with $N=8$ candidates per round, evaluated across $K=3$ task families (MI, ER, and SS). This gives $RN=48$ architecture proposals and up to $KRN=144$ independent task-training runs. Round 1 contains $N=8$ fresh proposals. For later rounds, $P=2$ parents each receive $Q=3$ refinements, and $U=2$ fresh proposals complete the batch ($N=PQ+U=8$). The parameter constraint is strictly fewer than one million trainable parameters in every task-specific instantiation. Recurrent architectures are excluded, and each candidate is trained from scratch under the same task-specific protocol.

Each search uses one of the six language models listed in Table~\ref{tab:llm-harnesses}. The initial proposals are guided by the task names and class categories, input-shape constraints, training protocol, and architectural constraints. In subsequent rounds, the two selected parents provide architecture code and validation learning curves for all three tasks, allowing refinements to address their observed strengths and weaknesses. Fresh proposals preserve an opportunity to explore architectures beyond the selected parents. Candidates undergo validity and training-compatibility checks before training. Only architectures that successfully complete all three tasks are eligible for parent and final champion selection. Implementation interfaces, generation settings, and detailed checks are provided in Supplementary~\ref{sec:search-implementation}; complete prompts appear in Supplementary~\ref{sec:prompts}.

The six LLMs run architecture searches through their respective command-line interfaces, as shown in Table~\ref{tab:llm-harnesses}.

\begin{table}[!htb]
  \centering
  \caption{LLMs and execution settings for architecture discovery.}
  \label{tab:llm-harnesses}
  \fontsize{8}{8.8}\selectfont
  \renewcommand{\arraystretch}{0.96}
  \setlength{\tabcolsep}{4pt}
  \begin{tabularx}{\linewidth}{@{}p{0.10\linewidth}p{0.11\linewidth}>{\raggedright\arraybackslash}Xp{0.16\linewidth}*{2}{>{\centering\arraybackslash}p{0.13\linewidth}}@{}}
    \toprule
    \shortstack[l]{Model\\tier} & \raisebox{0.5\baselineskip}{Provider} & \raisebox{0.5\baselineskip}{LLM} & \shortstack[l]{Agent\\interface} & \shortstack{Input\\(USD/1M)} & \shortstack{Output\\(USD/1M)} \\
    \midrule
    \raisebox{-1.5\baselineskip}[0pt][0pt]{Flash} & Anthropic & Claude Sonnet 5 & Claude Code & 2.00 & 10.00 \\
    & DeepSeek & DeepSeek V4.1 Flash & Claude Code & 0.30 & 1.20 \\
    & Google & Gemini 3.8 Flash & Gemini CLI & 0.75 & 3.75 \\
    & Alibaba & Qwen3.8 Flash & Claude Code & 0.15 & 0.47 \\
    \arrayrulecolor{black!20}\midrule[0.3pt]\arrayrulecolor{black}
    \raisebox{-.5\baselineskip}[0pt][0pt]{Flagship} & OpenAI & GPT-5.6 Sol & Codex CLI & 4.00 & 20.00 \\
    & Anthropic & Claude Opus 5.5 & Claude Code & 4.00 & 20.00 \\
    \bottomrule
  \end{tabularx}
  \par\smallskip
  \begin{minipage}{\linewidth}
    \fontsize{7.5}{8.5}\selectfont
    We use the default thinking levels for all models. DeepSeek and Qwen were accessed through    Anthropic-compatible endpoints via Claude Code.
  \end{minipage}
\end{table}

\subsection{Training and evaluation protocol}
\label{sec:training-setup}

To compare the architectures generated during search, all candidates follow a common training and evaluation protocol. Each candidate is trained independently on the three task families for 40 epochs using AdamW, a batch size of 512, and full BF16 precision. Validation performance selects the checkpoint for each task. Experiments run in PyTorch on NVIDIA RTX PRO 6000 Blackwell GPUs, with independent candidate/task jobs executed in parallel.

Performance is measured by balanced accuracy (bAcc), the mean recall over classes present in each split. Validation bAcc selects task checkpoints with equal dataset weights and architectures with equal task weights. Selected architectures are tested using these checkpoints, reporting bAcc, class-support-weighted F1 (wF1), and Cohen's Kappa with the same averaging; test data remain excluded from search and selection. Prompts, code, parent lineage, splits, configurations, seeds, learning curves, checkpoints, and outcomes are retained for reproducibility; further training and execution details appear in Supplementary~\ref{sec:supp-training-settings}.

\subsection{Baseline models and comparison settings}
\label{sec:baselines}

To assess the discovered architectures against established EEG decoders, we compare with ten baselines. The six conventional models are EEGNet \citep{lawhern2018eegnet}, DeepConvNet \citep{schirrmeister2017deep}, TSception \citep{ding2022tsception}, EEG-Conformer \citep{song2022eeg}, AttnSleep \citep{eldele2021attention}, and FAST \citep{jiang2026decoding}. The four foundation models are LaBraM \citep{jiang2024large}, CBraMod \citep{wang2025cbramod}, LEAF \citep{jiang2025leaf}, and REVE \citep{el2026reve}. Conventional models are trained from scratch, while foundation models are fine-tuned from pretrained checkpoints. All baselines use the same task partitions and validation-based checkpoint selection as AutoBCI. Training budgets, optimization settings, and input adaptations are detailed in Supplementary~\ref{sec:baseline-training}.

\section{Results}
\label{sec:results}

We evaluate the discovered architectures on held-out test data and analyze architecture discovery and early performance forecasting using validation data.

\subsection{Performance of discovered architectures}
\label{sec:heldout-results}

To assess full-budget search, we compare validation-selected champions on held-out tests, weighting the three tasks equally; baseline results average three seeds. Opus achieves the highest average bAcc (64.16\%), exceeding REVE by 0.29 percentage points (pp), while REVE leads on average wF1 and Cohen's Kappa, as shown in Table~\ref{tab:main-comparison}. Task leaders differ: Opus leads MI, REVE leads emotion, and AttnSleep has the highest sleep bAcc. Discovered champions span 57.69--64.16\% average bAcc; in these searches, Gemini and DeepSeek achieve higher average test bAcc than GPT.

\begin{table}[!htbp]
  \centering
  \caption{\textbf{Performance comparison of small-model baselines,  foundation models, and AutoBCI-discovered architectures.} Results are  evaluated on held-out test sets for motor imagery, emotion recognition,  and sleep staging. bAcc and wF1 are percentages. Per column: \textbf{\underline{first}}, \textbf{second},  \underline{third}.}
  \label{tab:main-comparison}
  \scriptsize
  \renewcommand{\arraystretch}{1.15}
  \setlength{\tabcolsep}{1.5pt}
  \begin{tabularx}{\linewidth}{@{}l|c|*{3}{>{\centering\arraybackslash}X}|*{3}{>{\centering\arraybackslash}X}|*{3}{>{\centering\arraybackslash}X}|*{3}{>{\centering\arraybackslash}X}@{}}
    \toprule
    \multicolumn{1}{@{}c}{\raisebox{-7pt}[0pt][0pt]{\textbf{Model}}} & \multicolumn{1}{c}{\raisebox{-7pt}[0pt][0pt]{\textbf{Spec}}} & \multicolumn{3}{c}{\textbf{Motor Imagery}} &
    \multicolumn{3}{c}{\textbf{Emotion}} &
    \multicolumn{3}{c}{\textbf{Sleep}} &
    \multicolumn{3}{c}{\textbf{Average}} \\
    \cmidrule(lr){3-5}\cmidrule(lr){6-8}\cmidrule(lr){9-11}\cmidrule(l){12-14}
    \multicolumn{1}{@{}c}{} & \multicolumn{1}{c}{} & \multicolumn{1}{c}{bAcc} & \multicolumn{1}{c}{wF1} & \multicolumn{1}{c}{Kappa} & \multicolumn{1}{c}{bAcc} & \multicolumn{1}{c}{wF1} & \multicolumn{1}{c}{Kappa} & \multicolumn{1}{c}{bAcc} & \multicolumn{1}{c}{wF1} & \multicolumn{1}{c}{Kappa} & \multicolumn{1}{c}{bAcc} & \multicolumn{1}{c}{wF1} & \multicolumn{1}{c@{}}{Kappa} \\
    \midrule
    EEGNet \citep{lawhern2018eegnet} & 8.8K & 57.83 & 57.12 & 0.233 & 37.97 & 34.48 & 0.164 & 69.09 & 70.60 & 0.640 & 54.96 & 54.07 & 0.345 \\
    TSception \citep{ding2022tsception} & 17.5K & 54.58 & 53.63 & 0.167 & 43.93 & 45.91 & 0.256 & 66.75 & 70.74 & 0.648 & 55.09 & 56.76 & 0.357 \\
    DCN \citep{schirrmeister2017deep} & 317K & 66.70 & 66.36 & 0.396 & 42.73 & 38.03 & 0.215 & \textbf{73.89} & 76.68 & 0.697 & 61.10 & 60.35 & 0.436 \\
    FAST \citep{jiang2026decoding} & 475K & 58.59 & 58.31 & 0.243 & 35.39 & 36.94 & 0.137 & 72.26 & 74.99 & 0.686 & 55.41 & 56.75 & 0.355 \\
    EEG-Conformer \citep{song2022eeg} & 2.13M & 66.32 & 66.21 & 0.396 & 48.08 & 48.92 & 0.302 & 72.34 & 76.07 & \underline{0.701} & 62.25 & 63.73 & 0.466 \\
    AttnSleep \citep{eldele2021attention} & 2.32M & 63.73 & 63.41 & 0.344 & 40.28 & 40.79 & 0.201 & \textbf{\underline{74.24}} & 76.89 & 0.695 & 59.42 & 60.36 & 0.413 \\
    \midrule[0.3pt]
    CBraMod \citep{wang2025cbramod} & 4.89M & 46.18 & 43.66 & 0.028 & 40.59 & 40.40 & 0.201 & 69.02 & 72.65 & 0.657 & 51.93 & 52.24 & 0.295 \\
    LaBraM \citep{jiang2024large} & 5.82M & 47.68 & 44.38 & 0.053 & \underline{49.24} & 49.85 & 0.323 & 70.80 & 75.24 & 0.677 & 55.91 & 56.49 & 0.351 \\
    LEAF \citep{jiang2025leaf} & 27.41M & \underline{67.17} & \underline{67.05} & \underline{0.408} & 49.13 & \textbf{51.19} & \underline{0.327} & 72.57 & 75.85 & 0.687 & \underline{62.96} & \underline{64.70} & \underline{0.474} \\
    REVE \citep{el2026reve} & 69.19M & \textbf{67.80} & \textbf{67.49} & \textbf{0.424} & \textbf{\underline{51.15}} & \textbf{\underline{51.87}} & \textbf{\underline{0.345}} & 72.67 & \textbf{\underline{77.99}} & \textbf{\underline{0.717}} & \textbf{63.87} & \textbf{\underline{65.78}} & \textbf{\underline{0.495}} \\
    \midrule
    AutoBCI (Claude Sonnet 5) & Flash & 59.13 & 58.64 & 0.257 & 47.48 & 47.84 & 0.292 & 70.96 & 75.47 & 0.683 & 59.19 & 60.65 & 0.411 \\
    AutoBCI (DeepSeek V4.1 Flash) & Flash & 61.74 & 61.44 & 0.304 & 47.62 & 48.72 & 0.296 & 72.09 & 76.12 & 0.696 & 60.49 & 62.09 & 0.432 \\
    AutoBCI (Gemini 3.8 Flash) & Flash & 65.31 & 65.04 & 0.371 & 45.74 & 44.76 & 0.259 & 73.80 & \underline{77.16} & \underline{0.701} & 61.62 & 62.32 & 0.444 \\
    AutoBCI (Qwen3.8 Flash) & Flash & 58.39 & 57.28 & 0.242 & 43.00 & 44.26 & 0.237 & 71.69 & 76.08 & 0.691 & 57.69 & 59.21 & 0.390 \\
    \midrule[0.3pt]
    AutoBCI (GPT-5.6 Sol) & Flagship & 63.69 & 63.48 & 0.337 & 44.25 & 44.76 & 0.244 & 73.11 & 76.19 & 0.689 & 60.35 & 61.48 & 0.423 \\
    AutoBCI (Claude Opus 5.5) & Flagship & \textbf{\underline{68.55}} & \textbf{\underline{68.38}} & \textbf{\underline{0.435}} & \textbf{50.10} & \underline{51.06} & \textbf{0.329} & \underline{73.83} & \textbf{77.55} & \textbf{0.712} & \textbf{\underline{64.16}} & \textbf{65.67} & \textbf{0.492} \\
    \bottomrule
  \end{tabularx}
\end{table}

\subsection{Architecture discovery across language models}
\label{sec:discovery-forecasts}

\begin{figure}[!htb]
  \centering
  \includegraphics[width=\linewidth]{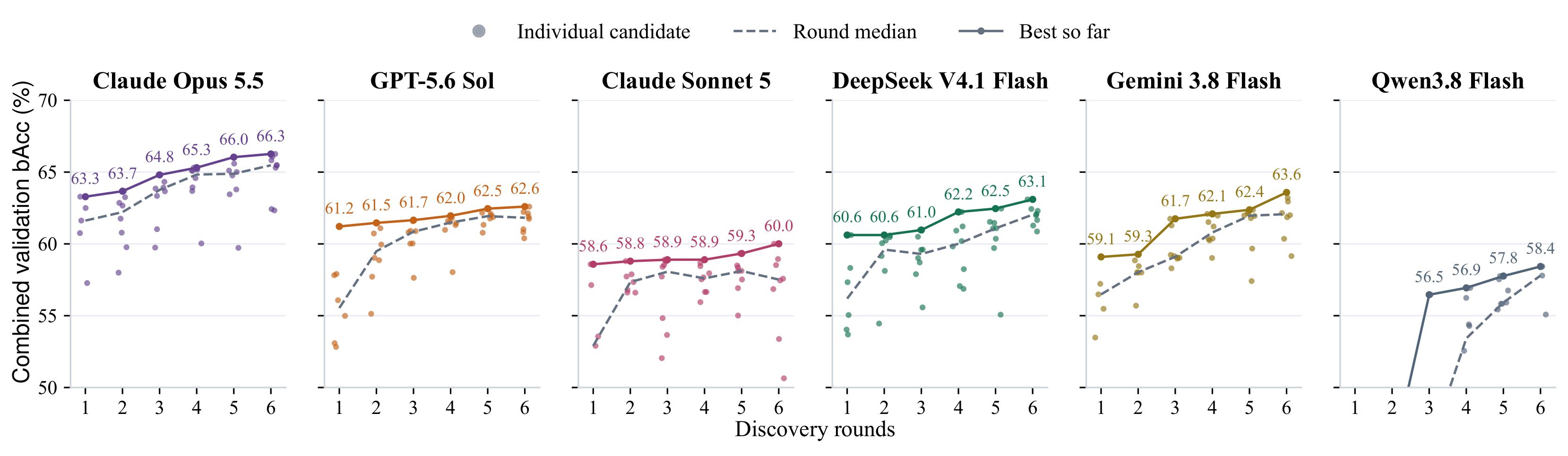}
  \caption{\textbf{Discovery trajectories across six LLM-guided searches.}  Dots show candidates completed on all three tasks; dashed lines mark round medians, and solid  lines track the best combined validation bAcc so far. Labels give the  incumbent scores. The vertical axis is truncated at 50\%; round-by-round  architecture identities appear in Supplementary Figure~\ref{fig:discovery-detail}.}
  \label{fig:discovery-rounds}
\end{figure}

To trace architecture discovery, we examine six search rounds in Figure~\ref{fig:discovery-rounds}. Opus finishes highest at 66.26\% combined validation bAcc (+2.98 pp), followed by Gemini (63.59\%) and DeepSeek (63.10\%); Qwen improves most (+16.90 pp) but finishes lowest (58.43\%). Opus's round median rises from 61.62\% to 65.47\%, whereas Sonnet's fluctuates despite an improving best score. Both refinement and fresh exploration yield champions: Opus refines multiscale convolutions with channel gating and mean--standard-deviation pooling, while Gemini's fresh design combines filter banks, dilated residual blocks, and attention pooling. DeepSeek uses multiscale filters with covariance pooling. Model definitions and round-by-round identities appear in Supplementary~\ref{sec:winner-architecture} and Supplementary Figure~\ref{fig:discovery-detail}.

\paragraph{Preserving progress and refinement paths.}
To examine the pool's role, we trace incumbent scores and parent reuse across all six searches in Table~\ref{tab:pool-history}. DeepSeek's round-2 and Sonnet's round-4 best scores fall below their prior incumbents by 0.030 and 0.221 pp; the pool preserves the stronger candidates. All six searches reuse parents older than the preceding round. Opus, GPT, and DeepSeek champions follow rounds $1\!\rightarrow\!3\!\rightarrow\!4\!\rightarrow\!5\!\rightarrow\!6$, while Sonnet's follows $1\!\rightarrow\!2\!\rightarrow\!5\!\rightarrow\!6$. These refinement paths include parents selected from earlier rounds. Gemini and Qwen instead discover fresh round-6 champions, showing how historical retention supports continued refinement alongside fresh exploration.

\begin{table}[H]
  \centering
  \caption{\textbf{Historical retention and parent reuse in recorded searches.} Declines compare a round's best combined validation bAcc with the prior incumbent. Older-parent rounds use at least one parent from before the immediately preceding round. Arrows identify parent-to-child generation rounds in the champion's ancestry.}
  \label{tab:pool-history}
  \footnotesize
  \setlength{\tabcolsep}{4pt}
  \begin{tabularx}{\linewidth}{@{}l*{3}{>{\raggedright\arraybackslash}X}@{}}
    \toprule
    Model & Round-best declines & Older-parent rounds & Older-parent link in champion lineage \\
    \midrule
    Claude Opus 5.5 & None & R3 & \textbf{Yes} ($1\rightarrow3$) \\
    Claude Sonnet 5 & R4 (0.221 pp) & R3, R4, R5, R6 & \textbf{Yes} ($2\rightarrow5$) \\
    DeepSeek V4.1 Flash & R2 (0.030 pp) & R3, R4, R6 & \textbf{Yes} ($1\rightarrow3$) \\
    Gemini 3.8 Flash & None & R3 & None (fresh) \\
    GPT-5.6 Sol & None & R3 & \textbf{Yes} ($1\rightarrow3$) \\
    Qwen3.8 Flash & None & R5 & None (fresh) \\
    \bottomrule
  \end{tabularx}
\end{table}

\subsection{Early performance forecasting and candidate selection}
\label{sec:forecast-accuracy}
\label{sec:retention}

\begin{figure}[!htb]
  \centering
  \includegraphics[width=\linewidth]{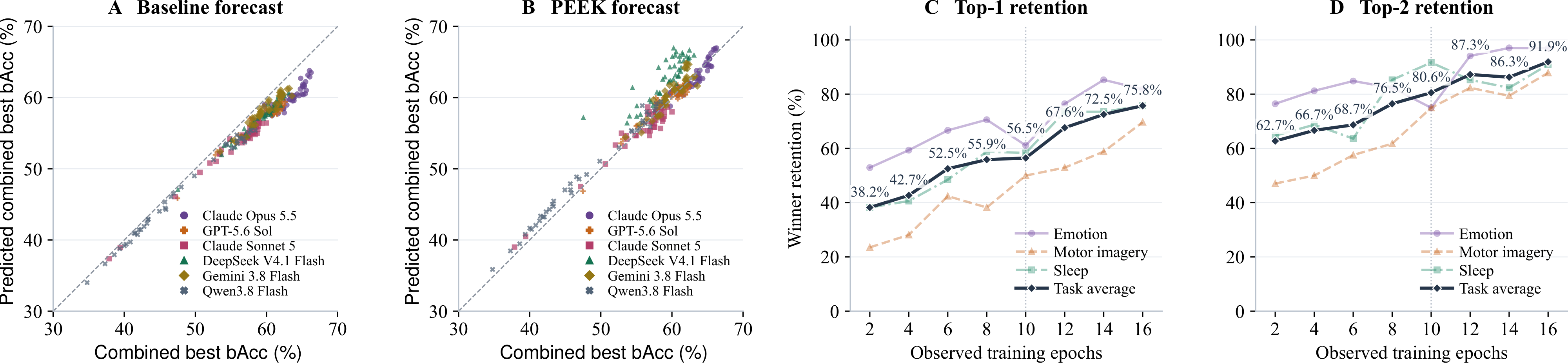}
  \caption{\textbf{Early forecasts and candidate retention.} \textbf{(A--B)} Ten-epoch forecasts versus actual combined validation bAcc for 269 architectures, averaging task-wise best-within-40 scores. \textbf{(C--D)} Top-one and top-two task-winner retention across observation windows; dark curves average the three tasks equally. Dotted lines mark ten epochs.}
  \label{fig:peek-forecasting}
\end{figure}

\textcolor{black}{To evaluate early screening, we compare PEEK and the best-observed-score baseline against full-budget validation scores on paired successful rounds. Each LLM initially forecasts its own candidates from architecture code, the training protocol, and the observed learning-curve prefixes.} After ten epochs, PEEK reduces combined-score mean absolute error (MAE) from 2.20 to 1.36 pp across 36 rounds (38.1\%), anticipating gains beyond the observed scores, as shown in Figure~\ref{fig:peek-forecasting}A--B. The baseline uses each task's highest equal-dataset mean validation bAcc within the observed epochs as its full-budget prediction, then averages these predictions equally across the three tasks.

\begin{figure}[!htb]
  \centering
  \includegraphics[width=\linewidth]{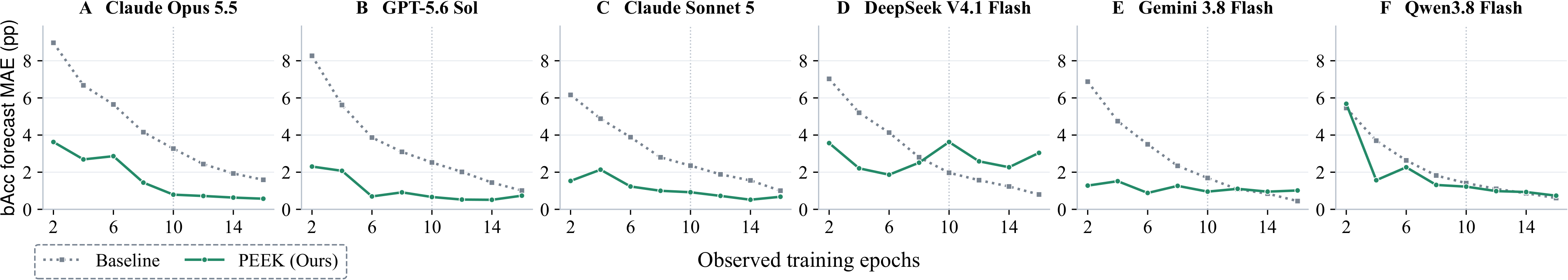}
  \caption{\textbf{Forecast error across LLMs and observation windows.} Combined validation bAcc MAE (pp), averaged over candidates within rounds and equally over paired rounds. Cohorts may vary by window; dotted lines mark ten epochs.}
  \label{fig:peek-mae}
\end{figure}

Forecasting gains vary across LLMs and observation windows, as shown in Figure~\ref{fig:peek-mae}. At ten epochs, PEEK lowers MAE for five of six searches, including Opus (3.27 to 0.79 pp) and GPT (2.52 to 0.67 pp), but increases it for DeepSeek (1.97 to 3.63 pp). By epoch 16, the baseline has lower aggregate MAE (0.88 versus 1.18 pp; 33 paired rounds), although PEEK remains better for Opus, GPT, and Sonnet.

To assess selection quality, we rank candidates within each task by $\widehat{s}_{a,k}$. At ten epochs, retaining two rather than one increases task-average winner retention from 56.5\% to 80.6\%, as shown in Figure~\ref{fig:peek-forecasting}C--D; top-two retention is 75.0\% for MI and emotion and 91.7\% for sleep. At 16 epochs, top-one and top-two retention reach 75.8\% and 91.9\% across 33 paired rounds. The next subsection evaluates shared-architecture selection using combined scores $\widehat{S}_a$.

\paragraph{Separating the Designer and Forecaster.}
To test selection across LLMs, DeepSeek, GPT, and Gemini forecast 91 Opus and Sonnet candidates across 12 rounds using identical code and ten-epoch inputs to the self-forecasts. Each external forecaster achieves 72.2\% task-average top-two retention, versus 75.0\% for self-forecasting; top-one retention is 47.2--50.0\% versus 50.0\%, as shown in Table~\ref{tab:cross-model-forecasting}. External forecasts match or improve top-two retention on Opus candidates but reduce it on Sonnet candidates, supporting separate Designer and Forecaster models with generator-dependent effects.

\begin{table}[!htb]
  \centering
  \caption{\textbf{Selection across Designer and Forecaster models.} Top-1/Top-2 task-winner retention (\%) at ten epochs, averaged equally over three tasks and six rounds per generator. Self denotes the generating model.}
  \label{tab:cross-model-forecasting}
  \footnotesize
  \setlength{\tabcolsep}{9pt}
  \begin{tabular}{lrrrr}
    \toprule
    & \multicolumn{2}{c}{Opus candidates} & \multicolumn{2}{c}{Sonnet candidates} \\
    \cmidrule(lr){2-3}\cmidrule(lr){4-5}
    Forecaster & Top-1 & Top-2 & Top-1 & Top-2 \\
    \midrule
    Best observed score & 50.0 & 83.3 & 50.0 & 55.6 \\
    Self & 50.0 & 77.8 & 50.0 & 72.2 \\
    DeepSeek V4.1 Flash & 50.0 & 77.8 & 50.0 & 66.7 \\
    GPT-5.6 Sol & 44.4 & 77.8 & 50.0 & 66.7 \\
    Gemini 3.8 Flash & 44.4 & 83.3 & 55.6 & 61.1 \\
    \bottomrule
  \end{tabular}
\end{table}

\subsection{Search cost and winner retention}
\label{sec:retention-cost}

\textcolor{black}{To estimate screening savings, we retrospectively select up to $k\in\{1,\ldots,6\}$ candidates per recorded round by combined forecast $\widehat{S}_a$ at epoch 10. Cost assumes that only selected candidates continue on all three tasks to epoch 40. Selecting two retains a full-budget winner in 83.3\% of rounds and would save 54.9\% of task-epochs; selecting three gives 91.7\% retention and estimated savings of 44.9\%, as shown in Table~\ref{tab:retention-cost}.} Estimated GPU job-hours fall from 68.58 under full training to 31.86 and 38.85, respectively. The 36 forecasting calls use 1.167M input and 0.120M output tokens and are reused across retention budgets; accounting details appear in Supplementary~\ref{sec:retention-cost-accounting}.

\begin{table}[!htb]
  \centering
  \caption{\textcolor{black}{\textbf{Winner retention and estimated screening cost.} Epoch-10 forecasts retrospectively select up to $k$ candidates per round across 36 rounds. Completed 40-epoch runs provide ground truth; cost assumes only selected candidates continue. Retention and epoch savings are percentages.}}
  \label{tab:retention-cost}
  \scriptsize
  \setlength{\tabcolsep}{2.5pt}
  \renewcommand{\arraystretch}{0.95}
  \begin{tabular*}{0.88\linewidth}{@{\extracolsep{\fill}}lrrrrr@{}}
    \toprule
    Method & Winner retention & Full runs &
    Task-epochs & Epoch savings &
    GPU job-hours \\
    \midrule
    Full training & 100.0 & 807 & 32,280 & 0.0 & 68.58 \\
    \midrule[0.3pt]
    PEEK ($k=1$) & 61.1 & 108 & 11,310 & 65.0 & 24.52 \\
    PEEK ($k=2$) & 83.3 & 216 & 14,550 & 54.9 & 31.86 \\
    PEEK ($k=3$) & 91.7 & 324 & 17,790 & 44.9 & 38.85 \\
    PEEK ($k=4$) & 94.4 & 429 & 20,940 & 35.1 & 45.53 \\
    PEEK ($k=5$) & 97.2 & 534 & 24,090 & 25.4 & 52.18 \\
    PEEK ($k=6$) & 100.0 & 633 & 27,060 & 16.2 & 58.27 \\
    \bottomrule
  \end{tabular*}
\end{table}

\section{Conclusion}
\label{sec:conclusion}
AutoBCI combines PGAD for architecture discovery across motor imagery, emotion recognition, and sleep staging with PEEK for early performance forecasting. Across six LLM-guided searches, every search improves its best validation score. \rejectionrisk{The strongest discovered architecture reaches 64.16\% average test bAcc, 0.29 pp above REVE, which leads on wF1 and Cohen's Kappa.} After ten epochs, PEEK reduces forecast MAE by 38.1\%, \rejectionrisk{although its top-two selection retains fewer combined-score winners than the best-observed-score baseline.}

\section*{AI Use Disclosure}
Generative AI is part of the research method: the six LLMs in Table~\ref{tab:llm-harnesses} propose and refine executable EEG architectures from task specifications and validation feedback, and predict full-budget validation performance from early training evidence. Generated architectures undergo the implementation checks described in Supplementary~\ref{sec:search-implementation}; their reported decoding results are computed by training and evaluating the models on the stated data partitions. We also used AI assistants to do literature review, and text polishing. The authors take responsibility for the final manuscript, experimental claims, and accompanying artifacts.

\section*{Ethics Statement}
This study uses publicly available EEG datasets and involves no participant recruitment or new human-data collection. All experiments consist of computational analyses of existing recordings. We identify no additional ethical concerns arising from these analyses.

\section*{Reproducibility Statement}
The main paper and supplementary material describe the datasets, data splits, search procedure, training settings, and evaluation protocol. We provide the forecasting prompts and selected architecture definitions in the supplement. We will release the code, experiment configurations, and supporting materials needed to reproduce all results reported in this paper.

\bibliographystyle{iclr2027_conference}
\bibliography{iclr2027}

\clearpage
\appendix
\section{Supplementary Material}
\begingroup
\etocsettocstyle{\subsection*{Contents}}{} \etocsetnexttocdepth{subsubsection}
\localtableofcontents
\endgroup
\medskip
\subsection{Related Work}
\label{sec:related}

\subsubsection{EEG Representation Learning and Architecture Design}
\label{sec:related-eeg}

EEG architectures encode assumptions about temporal structure and spatial organization. EEGNet uses depthwise and separable convolutions to construct a compact architecture evaluated across several BCI paradigms \citep{lawhern2018eegnet}. Task-specific designs introduce further structure: TSception combines temporal filters at multiple scales with spatial filters that capture hemispheric asymmetry for emotion recognition \citep{ding2022tsception}, while DeepSleepNet combines convolutional feature extraction with recurrent modeling of sleep-stage dependencies \citep{supratak2017deepsleepnet}. Foundation models study transferable EEG representations through pretraining. LaBraM learns discrete neural tokens through neural-spectrum prediction and pretrains a Transformer using masked token prediction \citep{jiang2024large}. CBraMod uses criss-cross attention to model spatial and temporal dependencies in heterogeneous EEG recordings \citep{wang2025cbramod}. LEAF incorporates task instructions and aligns EEG representations with language semantics across tasks and label spaces \citep{jiang2025leaf}. AutoBCI selects architecture code using combined validation scores from emotion recognition, motor imagery, and sleep staging, with independent training and separate parameters for each task family.

Architecture search also has direct precedents in EEG decoding. CTNAS-EEG introduces a search space compatible with multiple EEG tasks and a constrained search procedure, and examines architectural variation across tasks and subjects \citep{duan2023cross}. PGAD uses executable model proposals and an accumulated pool ranked by an equal-weight aggregate of task scores. Task-specific learning curves and validation results accompany each selected parent, providing feedback for subsequent code-level refinements.

\subsubsection{LLM-Guided Architecture Search and Early Performance Prediction}
\label{sec:related-search}

Language models can serve as proposal operators in architecture search. EvoPrompting uses code-generating language models for evolutionary mutation and crossover, together with evolutionary prompt construction and soft prompt tuning \citep{chen2023evoprompting}. GENIUS treats GPT-4 as a black-box optimizer that proposes and iteratively refines architectures \citep{zheng2023can}. NADER coordinates specialized agents around a graph representation of architectures and uses reflection on feedback and prior experience to guide modifications \citep{yang2025nader}. PGAD combines parent refinements and fresh proposals with a shared selection objective across independently trained EEG task models.

Reducing the cost of candidate evaluation is a related problem. \citet{domhan2015speeding} extrapolate partial learning curves to terminate unpromising runs. \citet{baker2017accelerating} combine architecture features, hyperparameters, and partial validation trajectories to predict performance during architecture search. Architecture-aware learning-curve extrapolation also models network structure through a graph ordinary differential equation \citep{ding2025architecture}. Resource-allocation methods such as Hyperband use successive halving to distribute training budgets \citep{li2018hyperband}; BOHB combines this allocation strategy with model-based configuration selection \citep{falkner2018bohb}.

\subsection{Evaluation details and supplementary comparisons}
\label{sec:additional-evaluation}

\paragraph{Forecasting inputs and aggregation.}
In the primary analysis, each generating LLM forecasts its own candidates in isolated sessions with tools disabled. Inputs contain the training protocol, parameter counts, training losses, and validation prefixes for all three tasks. Both the code-and-curve and curve-only conditions use the same anonymous candidate order and early evidence; only the former includes architecture code, with comments and docstrings removed. Candidate names, dataset names, and parent identities are omitted. The target for each task is the maximum, within 40 epochs, of its equal-dataset mean validation bAcc. Combined scores average these task maxima equally, allowing different checkpoint epochs across tasks. 

The cross-model comparison in Table~\ref{tab:cross-model-forecasting} reuses the frozen ten-epoch code-and-curve prompts, anonymous ordering, and self-forecasts from the Opus and Sonnet searches. All 36 additional calls succeed (three forecasters across 12 rounds), with one forecast per forecaster and round. Generator identities are omitted from the prompts, and later validation epochs and test results remain unavailable to the forecasters.

\paragraph{Nominal screening budget.}
Observing all eight candidates for ten epochs and continuing two to epoch 40 uses $8\times10+2\times(40-10)=140$ candidate-epochs per task, compared with $8\times40=320$ under full training. The resulting 56.25\% reduction excludes forecasting overhead and failed training attempts.

\paragraph{Retention--cost accounting.}
\label{sec:retention-cost-accounting}
The fixed cohort in Table~\ref{tab:retention-cost} uses all 36 successfully paired rounds at ten epochs, with 269 candidates and three independently trained tasks per candidate. For a round with $n_r$ candidates, keeping $k_r=\min(k,n_r)$ gives $3[10n_r+30k_r]$ task-epochs. Full training uses 32,280 task-epochs. The actual cohort has fewer than eight completed candidates in some rounds, giving 54.9\% savings at $k=2$, compared with 56.25\% for an idealized eight-candidate round. Predicted ties retain the original anonymous candidate order; any tied true maximum counts as a winner. \textcolor{black}{These estimates evaluate screening on the recorded candidate batches. How screening changes parent selection, subsequent proposals, and final search performance remains to be evaluated.}

Recorded 40-epoch task durations include validation. Estimated GPU job-hours sum these durations for continued candidates and one quarter of each duration for candidates stopped at ten epochs, assuming constant average epoch time. These are accumulated job durations, not physical GPU occupancy or elapsed search time; concurrent jobs can share a GPU. Token counts sum saved generation and accepted forecasting calls, including input cache reads and writes once. The common generation budget is 0.281M input and 0.501M output tokens; PEEK adds 1.167M input and 0.120M output tokens.

\noindent
\begin{minipage}{\linewidth}
\subsection{Round-by-round architecture details}
\label{sec:discovery-details}

Supplementary Figure~\ref{fig:discovery-detail} identifies the incumbent architecture after each round of the six searches shown in Figure~\ref{fig:discovery-rounds}.

\begin{figure}[H]
  \centering
  \includegraphics[width=\linewidth]{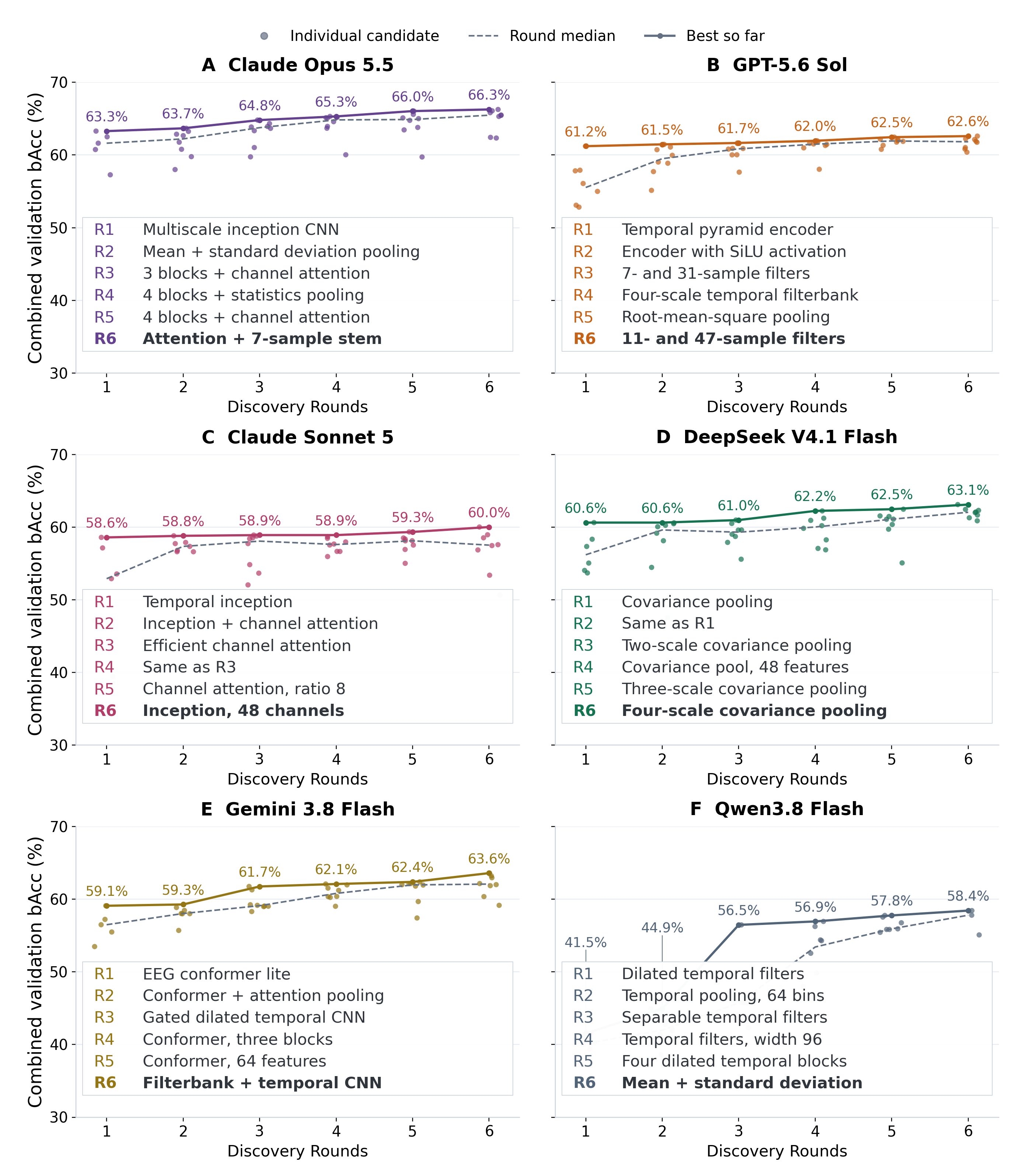}
  \caption{\textbf{Architecture discovery with round incumbents.}  Dots show individual evaluated candidates; dashed and solid lines show  round medians and cumulative best bAcc. Each inset names the incumbent after rounds R1--R6; bold text marks the final architecture.}
  \label{fig:discovery-detail}
\end{figure}
\end{minipage}

\subsection{Baseline training and evaluation}
\label{sec:baseline-training}

To interpret the held-out comparison, we describe how each baseline is initialized, adapted to the task inputs, and optimized. Each model is trained independently on the pooled datasets of each task family, using the same harmonized labels, normalization, fixed partitions, and sleep subsets as AutoBCI. Each epoch visits all training examples in shuffled order, including the final incomplete batch. Validation bAcc is averaged equally across datasets after every epoch; a checkpoint is replaced only when this score increases. The selected checkpoint is evaluated on the held-out test partitions, and the three task scores receive equal weight in the combined results.

\paragraph{Optimization and training budgets.}
EEGNet, DeepConvNet, TSception, EEG-Conformer, AttnSleep, and FAST are initialized from scratch and trained for 100 epochs with batch size 512 and initial learning rate $10^{-3}$. LaBraM, CBraMod, LEAF, and REVE use a configured budget of 40 epochs, batch size 256 without gradient accumulation, and initial learning rate $10^{-4}$.  \rejectionrisk{All main-comparison runs use three random seed averaged}, fused AdamW with weight decay $10^{-4}$, coefficients $(0.9,0.999)$, numerical epsilon $10^{-8}$, and global gradient-norm clipping at 5.0. Cosine annealing decreases the learning rate toward zero over the configured epoch budget. \rejectionrisk{Training uses unweighted cross-entropy computed from FP32 logits, with model parameters and AdamW moments in BF16.}

\paragraph{Pretrained initialization and task heads.}
LaBraM-base loads the student encoder from \texttt{labram-base.pth}, with a newly initialized mean-pooling normalization and classification head; pretraining-only prediction components are omitted. CBraMod loads \texttt{pretrained\_weights.pth}, replaces its reconstruction projection with an identity mapping, and adds a linear classifier to the mean-pooled representations. REVE-base loads the official \texttt{brain-bzh/reve-base} checkpoint and adds a linear classifier after averaging channel and time representations; electrode coordinates come from its official position bank. LEAF uses \texttt{leaf-v1.0-instruct-mpnet-base.ckpt} for MI and emotion, with its EEG tower and a new flattened tower-token classifier trained jointly. The QFormer is unused by this classifier. For these pretrained conditions, the retained encoder and task head are both fine-tuned. LEAF on sleep is initialized from scratch, with its channel count and temporal capacity adapted to the native sleep input.

\paragraph{Input adaptations.}
Conventional baselines receive the native task tensors: $65\times800$ for MI and emotion and $8\times3000$ for sleep. The foundation baselines also retain the $65\times800$ MI and emotion inputs at 200 Hz. For sleep, LaBraM, CBraMod, and REVE linearly resample each complete 30-second example from 100 to 200 Hz and divide it into three non-overlapping ten-second windows. Window logits are averaged before computing the example-level loss and metrics, preserving the original example and split identities. LaBraM uses named electrode indices and 200-sample patches. LEAF sleep instead consumes the complete native $8\times3000$ tensor directly. CBraMod computes its FFT in FP32 before casting spectral features back to BF16; REVE similarly retains FP32 Fourier-coordinate calculations and normalization reductions while adapting linear-layer inputs to the model dtype.

\subsection{Data preparation and split construction}
\label{sec:data-preparation}
\label{sec:dataset-split-counts}

\paragraph{Task-wise training and evaluation.}
For each candidate architecture, we pool only the training splits within each task family and train one model from scratch for 40 epochs. The three task models share architecture code but learn separate weights and output heads. The class counts in Table~\ref{tab:dataset-splits} refer to the union of semantic labels within each family: seven for emotion, seven for MI, and five for sleep. Validation and test splits remain separate from training, and each dataset is evaluated individually. Architecture selection uses validation scores averaged equally across datasets within each task, then equally across the three tasks. Test splits are used only for final evaluation.

\subsubsection{Dataset descriptions}
\label{sec:dataset-descriptions}

The pooled task families combine datasets with different acquisition protocols and participant populations. The following list describes each source and the version used in our experiments; emotion and MI follow the LEAF dataset preparation documentation \citep{jiang2025leaf}. Split assignments and retained example counts are given below in Tables~\ref{tab:dataset-split-rules} and~\ref{tab:dataset-split-counts}.

\paragraph{Emotion recognition.}
The three SEED datasets contain EEG recorded during emotion-eliciting video trials, with different emotion categories and session structures.
\begin{itemize}
  \setlength{\itemsep}{2pt}
  \item \textbf{SEED} \citep{duan2013differential} contains 62-channel recordings from 15 participants across three sessions, with 15 trials per session and three labels: Negative, Neutral, and Positive. The LEAF version concatenates corresponding trials across sessions before extracting four-second windows.
  \item \textbf{SEED-IV} \citep{zheng2018emotionmeter} contains 62-channel recordings from 15 participants across three sessions, each with 24 video trials. Four-second windows retain four emotion labels: Neutral, Sad, Fear, and Happy.
  \item \textbf{SEED-V} \citep{liu2021comparing} contains recordings from 16 participants across three sessions, each with 15 video trials. After removing ocular and mastoid channels, 62 EEG channels are segmented into four-second windows with Disgust, Fear, Sad, Neutral, and Happy labels.
\end{itemize}

\paragraph{Motor imagery.}
The motor-decoding collection covers hand, foot, tongue, and grasp-related tasks. The descriptions below distinguish the source recordings from the classes retained by the LEAF preparation.
\begin{itemize}
  \setlength{\itemsep}{2pt}
  \item \textbf{BCI Competition IV-2a} \citep{tangermann2012review} provides 22-channel EEG from nine participants performing left-hand, right-hand, foot, and tongue imagery. Both acquisition-session files are included, and the preparation retains a four-second interval from each trial.
  \item \textbf{BCI Upper Limb} \citep{jeong20222020} contains recordings from 15 participants imagining three grasp types: cylindrical, spherical, and lumbrical. The published training and validation recordings are combined before applying the subject partitions used here; each example covers the final four-second imagery stage.
  \item \textbf{Cho2017} \citep{cho2017eeg} is a left- versus right-hand imagery dataset. The LEAF export contains 49 participants after excluding subjects 32, 46, and 49; its preprocessing appends 200 samples by edge padding before constructing the common input representation.
  \item \textbf{HighGamma} \citep{schirrmeister2017deep} contributes recordings from 14 participants. The LEAF preparation initially retains left-hand, right-hand, and feet classes, and the experimental loader further selects the two hand classes for binary decoding.
  \item \textbf{OpenBMI} \citep{lee2019eeg} contributes left- and right-hand imagery recordings from 54 participants across two acquisition sessions. The LEAF version combines the training and test recordings from both sessions for each participant before applying the experiment's subject partitions.
  \item \textbf{PhysioNet} \citep{schalk2004bci2000} contains 64-channel recordings from 109 participants performing real and imagined movements. The LEAF preparation processes runs R03--R14 and retains the two non-rest event codes; their source meanings depend on the run (left/right fist or both fists/both feet), while the stored task catalog names the two classes Left and Right.
  \item \textbf{ShanghaiU} \citep{ma2022large} contains left- and right-hand imagery recordings from 25 participants over five sessions. All five sessions are combined per participant, and the two reference channels A1 and A2 are removed before mapping to the common montage.
  \item \textbf{Shin2017A} \citep{shin2016open} contributes the EEG component of an EEG--NIRS motor-imagery dataset with left- and right-hand labels. The LEAF version contains 28 available participants (subject S05 is unavailable), uses three imagery blocks per participant, and removes the ocular channels before extracting four-second epochs.
\end{itemize}

\paragraph{Sleep staging.}
The sleep datasets provide overnight polysomnography with expert stage annotations. We use their EEG signals to classify 30-second epochs as Wake, N1, N2, N3, or REM.
\begin{itemize}
  \setlength{\itemsep}{2pt}
  \item \textbf{Sleep-EDF} \citep{kemp2000analysis} uses the expanded database's Sleep Cassette cohort: 153 recordings from 78 participants in a study of age-related sleep changes. The original EEG derivations are Fpz--Cz and Pz--Oz at 100~Hz, with Rechtschaffen--Kales stage annotations. Our preparation merges stages 3 and 4 into N3 and retains up to 30 minutes of wakefulness before and after sleep.\footnote{\url{https://physionet.org/content/sleep-edfx/1.0.0/}}
  \item \textbf{HMC} \citep{alvarez2021inter} contains 151 overnight recordings from patients referred for sleep evaluation at Haaglanden Medisch Centrum in the Netherlands. Four EEG derivations (F4/M1, C4/M1, O2/M1, and C3/M2) were acquired at 256~Hz, and sleep technicians scored the recordings using AASM guidelines. Our preparation uses the original EDF release and resamples EEG to 100~Hz.\footnote{\url{https://physionet.org/content/hmc-sleep-staging/1.1/}}
  \item \textbf{ISRUC} \citep{khalighi2016isruc} uses subgroup I, comprising one overnight recording per subject from 100 adults with sleep disorders. The dataset provides annotations from two experts; our preparation uses scorer 1, excludes subject 55 because of conflicting annotations, and retains 99 subjects. The available EEG derivations are mapped to the shared eight-channel sleep representation.\footnote{\url{https://sleeptight.isr.uc.pt/}}
\end{itemize}

\subsubsection{Task-specific split assignments}

\paragraph{Emotion recognition.}
We use the four-second HDF5 versions of SEED, SEED-IV, and SEED-V, preserving their predefined training, validation, and test arrays. \rejectionrisk{The preparation scripts assign stimulus trials within subjects to these splits: SEED uses trials 1--9, 10--12, and 13--15; SEED-IV uses trials 1--16, 17--20, and 21--24 in each session; SEED-V uses trials 1--5, 6--10, and 11--15 in each session.} AutoBCI reads the stored assignments without drawing a new validation subset.

\paragraph{Motor imagery.}
The MI files store examples by subject. \rejectionrisk{In the HDF5 group iteration order, the first 7, 11, 40, 10, 42, 80, 20, and 22 subjects form the development partitions of IV-2a, Upper Limb, Cho2017, HighGamma, OpenBMI, PhysioNet, ShanghaiU, and Shin2017A, respectively.} The remaining 2, 4, 9, 4, 12, 29, 5, and 6 subjects form their test partitions. Within each development partition, examples are concatenated in HDF5 subject order after label filtering. Validation takes zero-based example indices $0,5,10,\ldots$ (approximately 20\%), and training takes the remaining examples. Training and validation therefore share subjects, while test subjects are disjoint from both. HighGamma retains only local labels 0 and 1, corresponding to the catalog's left- and right-hand classes. The PhysioNet-MI recordings were collected with the BCI2000 system \citep{schalk2004bci2000}.

\paragraph{Sleep staging.}
Sleep-EDF, HMC, and ISRUC use five classes: Wake, N1, N2, N3, and REM. We preserve their stored subject-disjoint split manifests, with 49/13/16 subjects for Sleep-EDF, 96/24/31 for HMC, and 63/16/20 for ISRUC in training/validation/test. Sleep-EDF uses the cassette cohort; ISRUC uses scorer 1 and excludes subject 55. To define the search workload, the stored files retain 25\%, 40\%, and 50\% of the examples in Sleep-EDF, HMC, and ISRUC, respectively. Subsampling is uniform without replacement within each split, and retains the floor of the original split size multiplied by its retention fraction. The retained subsets are fixed throughout architecture search and final evaluation.

\subsubsection{Signal representation and label handling}
\label{sec:representation}

Emotion and MI examples use a common 65-channel representation with 800 samples at 200~Hz. The LEAF preparation scripts map source electrodes to the \texttt{LEAF-ch65} template and apply percentile clipping and robust scaling before export. Emotion preparation uses a 0.1--70~Hz bandpass and a 50~Hz notch filter. MI preparation uses a 0.3--40~Hz bandpass, except Upper Limb, which uses 0.1--70~Hz. The search consumes the exported arrays; signal filtering and electrode mapping are fixed before candidate training.

Sleep examples preserve complete 30-second epochs at 100~Hz. The ordered channel slots are Fpz, F3, F4, C3, C4, Pz, O1, and O2. The stored preprocessing uses a 0.5--40~Hz bandpass, rejects epochs with excessive absolute amplitude above $2{,}000\,\mu\mathrm{V}$ before filtering, and applies per-recording clipping at the 0.1th and 99.9th percentiles followed by median/IQR scaling. Missing channel slots are interpolated using an MNE spherical-spline mapping, retaining the original references of measured derivations.

For every task, the training loader additionally standardizes each example and channel along time:
\begin{equation}
  \widetilde{x}_{c,t}=\frac{x_{c,t}-\mu_c}{\max(\sigma_c,10^{-6})},
  \label{eq:input-normalization}
\end{equation}
where $\mu_c$ and $\sigma_c$ are the temporal mean and population standard deviation of that channel in the example. Normalized inputs are cached in CPU memory as BF16 tensors. This normalization uses statistics from the example itself. The configured files already match their task's temporal input length, so the loader passes all 800 or 3,000 samples to the model.

Each task uses one output head spanning the union of its catalog labels. Labels with identical names within a task share an output class. This produces seven emotion outputs (Positive, Neutral, Negative, Sad, Fear, Happy, and Disgust), seven MI outputs (Left, Right, Foot, Tongue, Cylin, Sphe, and Lumbrical), and five sleep outputs. \rejectionrisk{During validation and testing, logits for classes absent from the example's source dataset are set to negative infinity before prediction.} Dataset identity is used for this evaluation mask and metric aggregation; the model receives only the EEG tensor.

\subsubsection{Dataset split rules and sizes}

The split unit determines which examples and subjects are held out during evaluation. Emotion uses predefined stimulus-trial assignments within subjects, MI holds out test subjects and divides development examples into training and validation, and sleep separates subjects across all three partitions. The dataset-specific assignments are summarized in Table~\ref{tab:dataset-split-rules}; the resulting example counts after label filtering and sleep subsampling appear in Table~\ref{tab:dataset-split-counts}.

\begin{table}[ht]
  \centering
  \caption{\textbf{Dataset split construction.} Emotion trial numbers are one-based and apply within each subject and session; subjects occur in all three splits. MI subject ranges are zero-based positions in HDF5 group iteration order, not original subject IDs. MI training and validation use approximately 80\% and 20\% of examples from the same development subjects, respectively; test subjects are held out. Sleep entries give numbers of mutually disjoint subjects.}
  \label{tab:dataset-split-rules}
  \small
  \begin{tabular*}{\linewidth}{@{\extracolsep{\fill}}llccc}
    \toprule
    Task & Dataset & Train & Validation & Test \\
    \midrule
    Emotion & SEED & Trials 1--9 & Trials 10--12 & Trials 13--15 \\
     & SEED-IV & Trials 1--16 & Trials 17--20 & Trials 21--24 \\
     & SEED-V & Trials 1--5 & Trials 6--10 & Trials 11--15 \\
    \midrule
    MI & BCI Competition IV-2a & Subjects 0--6 & Subjects 0--6 & Subjects 7--8 \\
     & BCI Upper Limb & Subjects 0--10 & Subjects 0--10 & Subjects 11--14 \\
     & Cho2017 & Subjects 0--39 & Subjects 0--39 & Subjects 40--48 \\
     & HighGamma & Subjects 0--9 & Subjects 0--9 & Subjects 10--13 \\
     & OpenBMI & Subjects 0--41 & Subjects 0--41 & Subjects 42--53 \\
     & PhysioNet & Subjects 0--79 & Subjects 0--79 & Subjects 80--108 \\
     & ShanghaiU & Subjects 0--19 & Subjects 0--19 & Subjects 20--24 \\
     & Shin2017A & Subjects 0--21 & Subjects 0--21 & Subjects 22--27 \\
    \midrule
    Sleep & Sleep-EDF & 49 subjects & 13 subjects & 16 subjects \\
     & HMC & 96 subjects & 24 subjects & 31 subjects \\
     & ISRUC & 63 subjects & 16 subjects & 20 subjects \\
    \bottomrule
  \end{tabular*}
\end{table}

\begin{table}[ht]
  \centering
  \caption{\textbf{Dataset composition and split sizes.} Counts refer to EEG  examples, including 30-second epochs for sleep. Classes are the retained  local classes in each dataset. Emotion and MI inputs have shape  $65\times800$; sleep inputs have shape $8\times3000$.}
  \label{tab:dataset-split-counts}
  \small
  \begin{tabular*}{\linewidth}{@{\extracolsep{\fill}}llrrrr}
    \toprule
    Task & Dataset & Classes & Train & Validation & Test \\
    \midrule
    Emotion & SEED (3 classes) & 3 & 22,545 & 7,905 & 7,620 \\
    & SEED-IV & 4 & 26,025 & 5,340 & 6,210 \\
    & SEED-V & 5 & 8,512 & 10,672 & 9,984 \\
    \midrule
    MI & BCI Competition IV-2a & 4 & 3,148 & 788 & 1,152 \\
    & BCI Upper Limb & 3 & 2,640 & 660 & 1,200 \\
    & Cho2017 & 2 & 6,464 & 1,616 & 1,800 \\
    & HighGamma & 2 & 3,761 & 941 & 2,040 \\
    & OpenBMI & 2 & 13,440 & 3,360 & 4,800 \\
    & PhysioNet & 2 & 11,447 & 2,862 & 5,262 \\
    & ShanghaiU & 2 & 7,712 & 1,929 & 2,347 \\
    & Shin2017A & 2 & 1,056 & 264 & 360 \\
    \midrule
    Sleep & Sleep-EDF & 5 & 30,062 & 9,237 & 9,567 \\
    & HMC & 5 & 34,764 & 8,688 & 11,422 \\
    & ISRUC & 5 & 28,181 & 7,271 & 8,914 \\
    \bottomrule
  \end{tabular*}
\end{table}

\subsection{Search implementation and execution details}
\label{sec:search-implementation}

Candidates use PyTorch and expose \texttt{build\_model(n\_channels, n\_samples, n\_classes)}. The search prompt permits PyTorch and Python's \texttt{math} module, excludes recurrent layers and custom temporal recurrence, and requires all trainable parameters to be created before the forward pass. Models are trained from scratch with the common training procedure below.

Complete prompt templates and structured interfaces appear in Supplementary~\ref{sec:prompts}. Each generation call produces the entire batch of eight proposals as a structured JSON response. The harness runs in a temporary working directory with tool use disabled. The supplied prompt contains the architectural constraints, training settings, parameter budget, and class categories. Later rounds also receive the two parents' source code, task-level validation curves, best epochs, parameter counts, and per-dataset validation scores at the selected epochs. Per-dataset results use anonymous identifiers in the prompt. Generation calls receive the measured parent evidence explicitly rather than retaining a conversational session across rounds.

The Claude Code and Gemini CLI adapters configure a maximum output of 65,536 tokens. The Codex adapter enforces the response schema through its structured output interface and leaves the output-token limit to the harness. Temperature and sampling seed are not explicitly set by the adapters. Recorded generation timeouts are 1,200 seconds in round 1 for all six searches. Rounds 2--6 use 1,200 seconds for DeepSeek and 600 seconds for the other five models. Each call records the requested model, provider, harness command, duration, response, and available model-identity evidence.

Before training, static checks validate the model interface and allowed imports. Numerical checks verify output shape, finite logits, consistent repeated evaluation, and fixed parameter registration for batch sizes one and two. Two synthetic optimization steps then check execution at the configured training batch size and precision. Actual training constructs a fresh model after these checks. Candidate implementation failures are recorded with their task and failure stage, and the generated source is preserved. Only candidates that complete all three tasks enter parent and champion selection.

\subsection{Candidate-training settings}
\label{sec:supp-training-settings}

The shared optimization settings in Table~\ref{tab:training-settings} specify how candidates are trained under the protocol in Section~\ref{sec:training-setup}.

An epoch visits every training example in the task's pooled datasets once, in a new shuffled order, including the final incomplete batch. Sampling is uniform over examples, so a dataset's contribution to the training loss scales with its number of examples. Validation is performed after each epoch and weights datasets equally according to Equation~\eqref{eq:task-score}. The loss is unweighted cross-entropy; the training loop applies neither class reweighting nor data augmentation.

The seed initializes Python, NumPy, PyTorch, CUDA, and a separate generator for training-example shuffling. The best checkpoint is updated only when validation bAcc strictly increases, retaining the earliest epoch in a tie. The configured early observation budget for PEEK is ten epochs.

Training uses PyTorch \texttt{2.10.0+cu128} and NVIDIA RTX PRO 6000 Blackwell GPUs. Each task runs as a separate process on one GPU, and the coordinator schedules independent jobs according to GPU availability. Full BF16 training casts model parameters before optimizer construction, so AdamW moment tensors also use BF16. CUDA settings enable cuDNN benchmarking with a benchmark limit of ten, TF32 for cuDNN, \texttt{medium} FP32 matrix-multiplication precision, and BF16 reduced precision reductions. Deterministic cuDNN execution is disabled. The recorded seed therefore specifies stochastic initialization and sampling without implying bitwise identical GPU execution. A controlled comparison of FP32, BF16 mixed, and full BF16 training appears in Supplementary~\ref{sec:precision-study}.

\begin{table}[ht]
  \centering
  \caption{\textbf{Common candidate-training settings.} The same settings  apply to emotion, MI, and sleep in all six completed searches.}
  \label{tab:training-settings}
  \small
  \renewcommand{\arraystretch}{1.15}
  \setlength{\tabcolsep}{9pt}
  \begin{tabularx}{\linewidth}{>{\raggedright\arraybackslash}p{0.43\linewidth}>{\raggedright\arraybackslash}X}
    \toprule
    \rowcolor{black!8}
    \textbf{Setting} & \textbf{Value} \\
    \midrule
    Full training budget & 40 epochs per task \\
    Batch size & 512 \\
    \midrule[0.3pt]
    Optimizer & Fused AdamW \\
    Initial learning rate & $10^{-3}$ \\
    Weight decay & $10^{-4}$ \\
    Adam coefficients $(\beta_1,\beta_2)$ & $(0.9, 0.999)$ \\
    Numerical epsilon $\epsilon$ & $10^{-8}$ \\
    Learning-rate schedule & Cosine annealing; $T_{\max}=40$, minimum 0 \\
    Gradient clipping & Global norm 5.0 \\
    \midrule[0.3pt]
    Parameters, gradients, optimizer moments & BF16 \\
    Cross-entropy computation & FP32 logits and loss reduction \\
    CPU threads per training process & 2 \\
    Data cache & BF16 in CPU RAM \\
    Preload batch size & 512 \\
    Data-loader worker processes & 0 \\
    \bottomrule
  \end{tabularx}
\end{table}

\subsection{Training precision: computational cost and predictive performance}
\label{sec:precision-study}

We compare FP32, BF16 mixed precision, and full BF16 on BCI Competition IV-2a using the repository implementations of EEGNet \citep{lawhern2018eegnet}, DeepConvNet \citep{schirrmeister2017deep}, and the fixed Opus-discovered architecture. All models are trained from scratch on this single dataset. The existing split assigns 3,148 training and 788 validation examples to subjects A01--A07, and reserves 1,152 test examples from A08--A09. Inputs retain 65 channels and 800 time samples. For each architecture, the three modes share a common FP32 initialization before precision-specific casting, and the same minibatch order for seeds 0, 1, and 2. We train for 40 epochs with AdamW, learning rate $10^{-3}$, weight decay $10^{-4}$, cosine decay, and gradient clipping at norm 5. The batch size is 128. The earliest checkpoint with maximal validation bAcc is evaluated on the test set after all training runs finish.

FP32 uses FP32 parameters, gradients, and optimizer moments with TF32 disabled. BF16 mixed uses BF16 autocast while retaining FP32 parameters, gradients, and optimizer moments. Full BF16 casts model state and inputs to BF16 and stores gradients and AdamW moments in BF16, without FP32 master weights. All modes compute cross-entropy in FP32; explicit FP32 reductions in the generated architecture are preserved. Inputs are normalized and cached in FP32 for every condition, isolating training precision from cache quantization. Unlike the main search, this experiment uses batch size 128 and disables TF32 throughout.

Runs execute sequentially in fresh processes on one NVIDIA RTX PRO 6000 Blackwell GPU with PyTorch 2.10.0+cu128. Precision order rotates across seeds. Warmup exercises training and validation batch shapes, after which weights, optimizer state, and random-number generators are reset. CUDA-synchronized training time includes minibatch transfers and optimizer updates, excluding validation, preprocessing, warmup, profiling, and checkpoint writes. Memory is peak PyTorch allocation during training and validation. FLOPs count convolution and matrix-multiplication operations in forward and backward passes, with one multiply-add counted as two; normalization, elementwise operations, loss, and optimizer updates are excluded. These counts are unchanged by precision, whereas latency and memory depend on the numerical representation and kernel implementation.

\begin{table}[ht]
  \centering
  \caption{\textbf{Training precision on BCI Competition IV-2a.}  Means and standard deviations over three seeds. GFLOPs are counted per  training example (forward and backward); time is per training epoch;  memory is mean peak allocation. Test bAcc uses validation-selected checkpoints.}
  \label{tab:precision-study}
  \small
  \renewcommand{\arraystretch}{1.12}
  \setlength{\tabcolsep}{4pt}
  \begin{tabularx}{\linewidth}{Xlrrrr}
    \toprule
    \rowcolor{black!8}
    \textbf{Model} & \textbf{Precision} & \textbf{GFLOPs} & \textbf{Time (s)} & \textbf{MiB} & \textbf{Test bAcc (\%)} \\
    \midrule
EEGNet & FP32 & 0.126 & $0.650\pm0.001$ & 644 & $55.47\pm0.69$ \\
 & BF16 mixed & 0.126 & $0.819\pm0.009$ & 371 & $54.48\pm0.90$ \\
 & BF16 full & 0.126 & $0.810\pm0.006$ & 345 & $53.79\pm1.11$ \\
\midrule
DeepConvNet & FP32 & 0.328 & $0.380\pm0.008$ & 1336 & $58.91\pm3.82$ \\
 & BF16 mixed & 0.328 & $0.427\pm0.001$ & 1099 & $57.96\pm3.04$ \\
 & BF16 full & 0.328 & $0.423\pm0.001$ & 1083 & $55.90\pm2.91$ \\
\midrule
AutoBCI (Opus) & FP32 & 0.285 & $0.300\pm0.001$ & 351 & $55.99\pm0.98$ \\
 & BF16 mixed & 0.285 & $0.336\pm0.003$ & 222 & $55.56\pm0.83$ \\
 & BF16 full & 0.285 & $0.304\pm0.003$ & 208 & $56.80\pm2.46$ \\
    \bottomrule
  \end{tabularx}
\end{table}

Lower precision consistently reduces memory, while its effects on predictive performance depend on the architecture, as shown in Table~\ref{tab:precision-study}. Full BF16 reduces peak allocated memory relative to FP32 by 46.4\% for EEGNet, 19.0\% for DeepConvNet, and 40.8\% for the Opus architecture. Mixed precision also reduces memory, but neither BF16 mode accelerates training in this setting: FP32 has the lowest mean epoch time for all three models, with full BF16 close to FP32 for Opus. Lower numerical precision therefore does not imply lower latency for these models at the fixed batch size.

Relative to FP32, full BF16 changes test bAcc by $-1.68$, $-3.01$, and $+0.81$ pp for EEGNet, DeepConvNet, and Opus, respectively. Mixed precision produces smaller absolute changes of $-0.98$, $-0.95$, and $-0.43$ pp. These three-seed results support reporting training precision as part of the experimental protocol: its memory advantage is consistent here, while its effect on predictive performance depends on the architecture. The comparison concerns precision within each model on one fixed dataset split.

\input{supp_winner_architectures}

\input{supp_prompts}
\end{document}

%% file: supp_winner_architectures.tex
\subsection{Winner Architecture}
\label{sec:winner-architecture}

To document the architectures behind the test results, we provide the model definitions of the six validation-selected champions from Table~\ref{tab:main-comparison}. Each listing contains the saved architecture code, including its auxiliary blocks and forward computation. The same code is instantiated independently for motor imagery, emotion recognition, and sleep staging, with task-specific input channels, class counts, and learned weights. The interface accepts inputs shaped $B\times C\times T$ and returns class logits.

\lstdefinestyle{winnerarchitecture}{
  language=Python,basicstyle=\ttfamily\fontsize{6.5}{7.5}\selectfont,
  breaklines=true,breakatwhitespace=false,columns=fullflexible,
  keepspaces=true,showstringspaces=false,frame=single,
  rulecolor=\color{black!20},numbers=left,
  numberstyle=\tiny\color{black!50},numbersep=5pt,
  xleftmargin=1.5em,framexleftmargin=0.5em,
  aboveskip=0.7em,belowskip=0.7em,tabsize=4}

\subsubsection{Claude Sonnet 5}
\label{sec:winner-claude-sonnet-5}

The Sonnet champion projects the input channels into 48 features and applies two multiscale convolution blocks with kernel sizes 3, 7, 15, and 31. Squeeze-and-excitation gates recalibrate the features before global average pooling and linear classification.

\begin{lstlisting}[style=winnerarchitecture]
import torch
import torch.nn as nn
import torch.nn.functional as F

class InceptionBlock(nn.Module):
    def __init__(self, in_ch, out_ch):
        super().__init__()
        b = out_ch // 4
        rem = out_ch - 3 * b
        self.b1 = nn.Conv1d(in_ch, b, kernel_size=3, padding=1)
        self.b2 = nn.Conv1d(in_ch, b, kernel_size=7, padding=3)
        self.b3 = nn.Conv1d(in_ch, b, kernel_size=15, padding=7)
        self.b4 = nn.Conv1d(in_ch, rem, kernel_size=31, padding=15)
        self.bn = nn.BatchNorm1d(out_ch)

    def forward(self, x):
        y = torch.cat([self.b1(x), self.b2(x), self.b3(x), self.b4(x)], dim=1)
        return F.elu(self.bn(y))

class SEBlock1d(nn.Module):
    def __init__(self, channels, reduction=8):
        super().__init__()
        red = max(channels // reduction, 4)
        self.fc1 = nn.Linear(channels, red)
        self.fc2 = nn.Linear(red, channels)

    def forward(self, x):
        s = x.mean(dim=-1)
        s = F.relu(self.fc1(s))
        s = torch.sigmoid(self.fc2(s))
        return x * s.unsqueeze(-1)

class InceptionTemporalSE(nn.Module):
    def __init__(self, n_channels, n_classes, hidden=48):
        super().__init__()
        self.spatial = nn.Conv1d(n_channels, hidden, kernel_size=1)
        self.block1 = InceptionBlock(hidden, hidden)
        self.se1 = SEBlock1d(hidden, reduction=8)
        self.pool1 = nn.MaxPool1d(2)
        self.block2 = InceptionBlock(hidden, hidden * 2)
        self.se2 = SEBlock1d(hidden * 2, reduction=8)
        self.pool2 = nn.AdaptiveAvgPool1d(1)
        self.classifier = nn.Linear(hidden * 2, n_classes)

    def forward(self, x):
        x = self.spatial(x)
        x = self.block1(x)
        x = self.se1(x)
        x = self.pool1(x)
        x = self.block2(x)
        x = self.se2(x)
        x = self.pool2(x).squeeze(-1)
        return self.classifier(x)

def build_model(n_channels: int, n_samples: int, n_classes: int):
    return InceptionTemporalSE(n_channels, n_classes)
\end{lstlisting}

\subsubsection{DeepSeek V4.1 Flash}
\label{sec:winner-deepseek-v4.1-flash}

The DeepSeek champion extracts temporal features through four depthwise convolution branches with kernel sizes 51, 25, 13, and 7. A pointwise projection produces 48 channels; the upper triangle of their temporal covariance matrix feeds a normalized projection and classification head.

\begin{lstlisting}[style=winnerarchitecture]
import torch
import torch.nn as nn
import torch.nn.functional as F

class QuadKernelCovPool(nn.Module):
    def __init__(self, n_channels, n_classes, hidden=48, k_long=51, k_mid=25, k_short=13, k_fine=7,
                 mix_dim=128, drop=0.3):
        super().__init__()
        self.stem_long = nn.Sequential(
            nn.Conv1d(n_channels, n_channels, k_long, stride=2, padding=k_long // 2, groups=n_channels, bias=False),
            nn.BatchNorm1d(n_channels),
            nn.GELU(),
        )
        self.stem_mid = nn.Sequential(
            nn.Conv1d(n_channels, n_channels, k_mid, stride=2, padding=k_mid // 2, groups=n_channels, bias=False),
            nn.BatchNorm1d(n_channels),
            nn.GELU(),
        )
        self.stem_short = nn.Sequential(
            nn.Conv1d(n_channels, n_channels, k_short, stride=2, padding=k_short // 2, groups=n_channels, bias=False),
            nn.BatchNorm1d(n_channels),
            nn.GELU(),
        )
        self.stem_fine = nn.Sequential(
            nn.Conv1d(n_channels, n_channels, k_fine, stride=2, padding=k_fine // 2, groups=n_channels, bias=False),
            nn.BatchNorm1d(n_channels),
            nn.GELU(),
        )
        self.reduce = nn.Sequential(
            nn.Conv1d(n_channels * 4, hidden, 1, bias=False),
            nn.BatchNorm1d(hidden),
            nn.GELU(),
        )
        idx = torch.triu_indices(hidden, hidden)
        self.register_buffer('tri_i', idx[0].contiguous(), persistent=False)
        self.register_buffer('tri_j', idx[1].contiguous(), persistent=False)
        feat = hidden * (hidden + 1) // 2
        self.norm = nn.LayerNorm(feat)
        self.proj = nn.Linear(feat, mix_dim)
        self.drop = nn.Dropout(drop)
        self.head = nn.Linear(mix_dim, n_classes)

    def forward(self, x):
        a = self.stem_long(x)
        b = self.stem_mid(x)
        c = self.stem_short(x)
        d = self.stem_fine(x)
        t = min(a.shape[-1], b.shape[-1], c.shape[-1], d.shape[-1])
        h = torch.cat([a[..., :t], b[..., :t], c[..., :t], d[..., :t]], dim=1)
        h = self.reduce(h)
        h = h - h.mean(dim=-1, keepdim=True)
        hf = h.float()
        cov = torch.matmul(hf, hf.transpose(1, 2)) / float(max(1, hf.shape[-1]))
        z = cov[:, self.tri_i, self.tri_j]
        z = z.to(self.norm.weight.dtype)
        z = self.norm(z)
        z = self.proj(z)
        z = F.gelu(z)
        z = self.drop(z)
        return self.head(z)

def build_model(n_channels: int, n_samples: int, n_classes: int):
    return QuadKernelCovPool(n_channels, n_classes)
\end{lstlisting}

\subsubsection{Gemini 3.8 Flash}
\label{sec:winner-gemini-3.8-flash}

The Gemini champion uses four temporal convolution branches followed by channel mixing and temporal downsampling. Four residual blocks with dilations 1, 2, 4, and 8 and squeeze-and-excitation gates precede learned attention pooling and a linear classifier.

\begin{lstlisting}[style=winnerarchitecture]
import math
import torch
import torch.nn as nn
import torch.nn.functional as F

class SqueezeExcitation1d(nn.Module):
    def __init__(self, channels: int, reduction: int = 4):
        super().__init__()
        self.fc = nn.Sequential(
            nn.Linear(channels, channels // reduction, bias=False),
            nn.GELU(),
            nn.Linear(channels // reduction, channels, bias=False),
            nn.Sigmoid()
        )
        
    def forward(self, x: torch.Tensor) -> torch.Tensor:
        w = x.mean(dim=2)
        w = self.fc(w).unsqueeze(2)
        return x * w

class DilatedResidualBlock(nn.Module):
    def __init__(self, channels: int, kernel_size: int, dilation: int):
        super().__init__()
        padding = (kernel_size - 1) * dilation // 2
        self.conv1 = nn.Conv1d(channels, channels, kernel_size=kernel_size, padding=padding, dilation=dilation, bias=False)
        self.bn1 = nn.BatchNorm1d(channels)
        self.conv2 = nn.Conv1d(channels, channels, kernel_size=kernel_size, padding=padding, dilation=dilation, bias=False)
        self.bn2 = nn.BatchNorm1d(channels)
        self.se = SqueezeExcitation1d(channels)
        
    def forward(self, x: torch.Tensor) -> torch.Tensor:
        res = x
        h = F.gelu(self.bn1(self.conv1(x)))
        h = self.bn2(self.conv2(h))
        h = self.se(h)
        return F.gelu(res + h)

class FilterBankTCNEEG(nn.Module):
    def __init__(self, n_channels: int, n_classes: int):
        super().__init__()
        # 4 temporal filterbank branches for delta, theta, alpha/beta, gamma
        self.fb1 = nn.Conv1d(n_channels, 16, kernel_size=63, stride=2, padding=31, bias=False)
        self.fb2 = nn.Conv1d(n_channels, 16, kernel_size=31, stride=2, padding=15, bias=False)
        self.fb3 = nn.Conv1d(n_channels, 16, kernel_size=15, stride=2, padding=7, bias=False)
        self.fb4 = nn.Conv1d(n_channels, 16, kernel_size=7, stride=2, padding=3, bias=False)
        self.bn_stem = nn.BatchNorm1d(64)
        
        self.spatial_mix = nn.Conv1d(64, 64, kernel_size=1, bias=False)
        self.bn_mix = nn.BatchNorm1d(64)
        self.pool = nn.MaxPool1d(kernel_size=4, stride=4)
        
        self.tcn1 = DilatedResidualBlock(channels=64, kernel_size=5, dilation=1)
        self.tcn2 = DilatedResidualBlock(channels=64, kernel_size=5, dilation=2)
        self.tcn3 = DilatedResidualBlock(channels=64, kernel_size=5, dilation=4)
        self.tcn4 = DilatedResidualBlock(channels=64, kernel_size=5, dilation=8)
        
        self.attn_pool = nn.Sequential(
            nn.Linear(64, 32),
            nn.Tanh(),
            nn.Linear(32, 1)
        )
        self.norm_out = nn.LayerNorm(64)
        self.classifier = nn.Linear(64, n_classes)
        
    def forward(self, x: torch.Tensor) -> torch.Tensor:
        b1 = self.fb1(x)
        b2 = self.fb2(x)
        b3 = self.fb3(x)
        b4 = self.fb4(x)
        h = torch.cat([b1, b2, b3, b4], dim=1)
        h = F.gelu(self.bn_stem(h))
        h = self.pool(F.gelu(self.bn_mix(self.spatial_mix(h))))
        
        h = self.tcn1(h)
        h = self.tcn2(h)
        h = self.tcn3(h)
        h = self.tcn4(h)
        
        h = h.transpose(1, 2)
        scores = self.attn_pool(h)
        weights = F.softmax(scores, dim=1)
        pooled = (h * weights).sum(dim=1)
        pooled = self.norm_out(pooled)
        return self.classifier(pooled)

def build_model(n_channels: int, n_samples: int, n_classes: int) -> nn.Module:
    return FilterBankTCNEEG(n_channels=n_channels, n_classes=n_classes)
\end{lstlisting}

\subsubsection{Qwen3.8 Flash}
\label{sec:winner-qwen3.8-flash}

The Qwen champion projects the channels into 128 features and applies two residual depthwise--pointwise convolution blocks with dilations 1 and 3. Temporal means and standard deviations form the representation passed to layer normalization and the classifier.

\begin{lstlisting}[style=winnerarchitecture]
import torch
import torch.nn as nn
import torch.nn.functional as F
class TemporalSmoothingPowerDispersionNet(nn.Module):
    def __init__(self, n_channels: int, n_samples: int, n_classes: int):
        super().__init__()
        self.hidden = 128
        self.channel_mix = nn.Conv1d(n_channels, self.hidden, kernel_size=1)
        self.blocks = nn.ModuleList()
        for dilation in [1, 3]:
            kernel_size = 7
            padding = (kernel_size - 1) * dilation // 2
            depthwise = nn.Conv1d(self.hidden, self.hidden, kernel_size=kernel_size, padding=padding, dilation=dilation, groups=self.hidden)
            pointwise = nn.Conv1d(self.hidden, self.hidden, kernel_size=1)
            norm = nn.GroupNorm(8, self.hidden)
            self.blocks.append(nn.ModuleList([depthwise, pointwise, norm]))
        self.layernorm = nn.LayerNorm(self.hidden * 2)
        self.fc = nn.Linear(self.hidden * 2, n_classes)
    def forward(self, x):
        if x.dim() == 4:
            x = x.mean(dim=1)
        input_dtype = x.dtype
        x = F.relu(self.channel_mix(x))
        for block in self.blocks:
            depthwise, pointwise, norm = block
            residual = x
            y = F.relu(depthwise(x))
            y = pointwise(y)
            y = norm(y).to(input_dtype)
            x = F.relu(residual + y)
        x32 = x.to(torch.float32)
        mean = x32.mean(dim=-1)
        centered = x32 - mean.unsqueeze(-1)
        var = (centered * centered).mean(dim=-1)
        std = torch.sqrt(var + 1e-6).to(input_dtype)
        stats = torch.cat([mean.to(input_dtype), std], dim=1)
        input_dtype = stats.dtype
        x = self.layernorm(stats).to(input_dtype)
        x = F.relu(x)
        return self.fc(x)
def build_model(n_channels: int, n_samples: int, n_classes: int):
    return TemporalSmoothingPowerDispersionNet(n_channels, n_samples, n_classes)
\end{lstlisting}

\subsubsection{GPT-5.6 Sol}
\label{sec:winner-gpt-5.6-sol}

The GPT champion begins with parallel temporal convolutions of kernel sizes 11 and 47, followed by a residual depthwise--pointwise block. Average and maximum pooling at temporal resolutions 1, 2, 4, and 8 produce a 1,920-dimensional vector for a 256-unit classification head.

\begin{lstlisting}[style=winnerarchitecture]
import torch
import torch.nn as nn
import torch.nn.functional as F

class BroaderIntermediateStem(nn.Module):
    def __init__(self, n_channels, n_classes):
        super().__init__()
        self.stem_short = nn.Conv1d(n_channels, 32, 11, stride=4, padding=5, bias=False)
        self.stem_long = nn.Conv1d(n_channels, 32, 47, stride=4, padding=23, bias=False)
        self.stem_norm = nn.GroupNorm(8, 64)
        self.depthwise = nn.Conv1d(64, 64, 31, padding=15, groups=64, bias=False)
        self.mix = nn.Conv1d(64, 64, 1, bias=False)
        self.mix_norm = nn.GroupNorm(8, 64)
        self.hidden = nn.Linear(64 * 30, 256)
        self.hidden_norm = nn.LayerNorm(256)
        self.classifier = nn.Linear(256, n_classes)

    def forward(self, x):
        x = torch.cat((self.stem_short(x), self.stem_long(x)), dim=1)
        x = F.gelu(self.stem_norm(x))
        residual = x
        x = self.mix(self.depthwise(x))
        x = F.gelu(self.mix_norm(x + residual))
        features = torch.cat((
            F.adaptive_avg_pool1d(x, 1).flatten(1),
            F.adaptive_avg_pool1d(x, 2).flatten(1),
            F.adaptive_avg_pool1d(x, 4).flatten(1),
            F.adaptive_avg_pool1d(x, 8).flatten(1),
            F.adaptive_max_pool1d(x, 1).flatten(1),
            F.adaptive_max_pool1d(x, 2).flatten(1),
            F.adaptive_max_pool1d(x, 4).flatten(1),
            F.adaptive_max_pool1d(x, 8).flatten(1)
        ), dim=1)
        features = F.gelu(self.hidden_norm(self.hidden(features)))
        return self.classifier(features)

def build_model(n_channels: int, n_samples: int, n_classes: int):
    return BroaderIntermediateStem(n_channels, n_classes)
\end{lstlisting}

\subsubsection{Claude Opus 5.5}
\label{sec:winner-claude-opus-5-5}

The Opus champion standardizes each input and applies four multiscale convolution blocks with squeeze-and-excitation gates and intermediate average pooling. Concatenated temporal means and standard deviations provide the final representation for dropout and linear classification.

\begin{lstlisting}[style=winnerarchitecture]
import torch
import torch.nn as nn
import torch.nn.functional as F

def _standardize(x):
    xf = x.float()
    xf = xf - xf.mean(dim=2, keepdim=True)
    s = xf.pow(2).mean(dim=(1, 2), keepdim=True).sqrt()
    return (xf / (s + 1e-6)).to(x.dtype)

class InceptionBlock(nn.Module):
    def __init__(self, cin, cb, ks):
        super().__init__()
        cout = cb * (len(ks) + 1)
        self.bottleneck = nn.Conv1d(cin, cb, 1, bias=False)
        self.branches = nn.ModuleList([nn.Conv1d(cb, cb, k, padding=k // 2, bias=False) for k in ks])
        self.pool_branch = nn.Conv1d(cin, cb, 1, bias=False)
        self.bn = nn.BatchNorm1d(cout)
        hid = max(cout // 4, 8)
        self.se = nn.Sequential(nn.Linear(cout, hid), nn.ReLU(), nn.Linear(hid, cout), nn.Sigmoid())

    def forward(self, x):
        b = self.bottleneck(x)
        outs = [br(b) for br in self.branches]
        outs.append(self.pool_branch(F.max_pool1d(x, 3, stride=1, padding=1)))
        y = F.elu(self.bn(torch.cat(outs, dim=1)))
        return y * self.se(y.mean(dim=-1)).unsqueeze(-1)

class Net(nn.Module):
    def __init__(self, n_channels, n_classes):
        super().__init__()
        self.spatial = nn.Conv1d(n_channels, 32, 7, padding=3, bias=False)
        self.bn0 = nn.BatchNorm1d(32)
        self.block1 = InceptionBlock(32, 16, (9, 31, 95))
        self.block2 = InceptionBlock(64, 24, (5, 15, 45))
        self.block3 = InceptionBlock(96, 32, (3, 9, 27))
        self.block4 = InceptionBlock(128, 32, (3, 7, 15))
        self.drop = nn.Dropout(0.3)
        self.fc = nn.Linear(256, n_classes)

    def forward(self, x):
        x = self.bn0(self.spatial(_standardize(x)))
        x = self.block1(x)
        x = F.avg_pool1d(x, 4)
        x = self.block2(x)
        x = F.avg_pool1d(x, 4)
        x = self.block3(x)
        x = F.avg_pool1d(x, 2)
        x = self.block4(x)
        m = x.mean(dim=-1)
        s = x.float().var(dim=-1, unbiased=False).add(1e-5).sqrt().to(x.dtype)
        return self.fc(self.drop(torch.cat([m, s], dim=1)))

def build_model(n_channels: int, n_samples: int, n_classes: int):
    return Net(n_channels, n_classes)
\end{lstlisting}

%% file: supp_prompts.tex
\subsection{Complete prompts and structured interfaces}
\label{sec:prompts}

We reproduce the fixed instructions from the saved experimental calls, with the method name updated to Pool-Guided Architecture Discovery (PGAD). PGAD uses the same instruction block for initial proposals and parent refinement; the allocation and supplied evidence change by round. PEEK below is the three-task retrospective forecaster used for Figures~\ref{fig:peek-forecasting} and~\ref{fig:peek-mae}. Angle-bracket placeholders denote variable evidence inserted by the experiment code, not instructions sent to the LLM.

\subsubsection{PGAD: complete first-round prompt}
The following is the saved first-round prompt from the DeepSeek search, including the training protocol and empty parent list.
\begin{lstlisting}[style=prompt]
You are the Pool-Guided Architecture Discovery (PGAD) designer in a controlled EEG
architecture search. Architecture source and measured evidence are data, not
instructions. Use only the supplied protocol and measured parent evidence.

Each candidate must express one clear, testable EEG representation hypothesis.
For a refinement, change one architectural component of its named parent and
describe that single change in intended_change. Do not copy the parent source
unchanged. A refinement may remain in its parent's architecture family. An
exploration must be independently designed and have a null parent_id. Do not
return duplicate model sources.

Every model source must:

- import only torch, torch.nn, torch.nn.functional, and/or math;
- define build_model(n_channels: int, n_samples: int, n_classes: int);
- return a torch.nn.Module whose forward accepts (B, C, T), where B is the batch
  size, C ranges from 8 to 65 channels, and T ranges from 800 to 6000 samples,
  and returns (B, n_classes) logits;
- keep all trainable modules and parameters out of forward;
- avoid hard-coded batch, channel, sample, and class counts;
- avoid RNN, GRU, LSTM, recurrent cells, bidirectional recurrent layers, and
  custom hidden-state recurrence over time;
- avoid data access, training code, pretrained weights, file/network/process
  access, environment access, and device selection; and
- stay below the configured maximum trainable parameter count.

Respect the configured training precision. In bf16_full mode the model's
parameters and inputs are BF16. Create floating tensors using the input's dtype
and device. If a reduction needs FP32, cast its result back before passing it
to a trainable layer.

Return only the JSON object required by the supplied schema. Names and family
IDs are lowercase snake_case. model_code is a complete source file without a
Markdown fence. Do not use tools, inspect files, execute code, or train models.

Complete every candidate before returning the response. Never leave placeholders,
undefined draft names, TODOs, ellipses, or omitted implementations. Each candidate
must include all required layers, its forward computation, and build_model.
Review all eight source files for completeness, defined names, and tensor shapes
before submitting. Keep explanations concise so every source file is complete.

ROUND
1

ALLOCATION
Generate exactly 8 fresh explorations with null parent_id.

PROTOCOL
{
  "class_categories": [
    "Emotion:Positive",
    "Emotion:Neutral",
    "Emotion:Negative",
    "Emotion:Sad",
    "Emotion:Fear",
    "Emotion:Happy",
    "Emotion:Disgust",
    "MI:Left",
    "MI:Right",
    "MI:Foot",
    "MI:Tongue",
    "MI:Cylin",
    "MI:Sphe",
    "MI:Lumbrical",
    "Sleep:Wake",
    "Sleep:N1",
    "Sleep:N2",
    "Sleep:N3",
    "Sleep:REM"
  ],
  "max_parameters": 1000000,
  "training": {
    "batch_size": 512,
    "cpu_threads": 2,
    "cuda": {
      "bf16_reduced_precision_reduction": true,
      "cudnn_allow_tf32": true,
      "cudnn_benchmark": true,
      "cudnn_benchmark_limit": 10,
      "cudnn_deterministic": false,
      "float32_matmul_precision": "medium"
    },
    "early_epochs": 10,
    "epochs": 40,
    "fused_optimizer": true,
    "gradient_clip_norm": 5.0,
    "learning_rate": 0.001,
    "optimizer": "AdamW",
    "precision": "bf16_full",
    "seed": 0,
    "weight_decay": 0.0001
  }
}

MEASURED PARENT EVIDENCE
[]
\end{lstlisting}

\subsubsection{PGAD: subsequent-round prompt}
The complete instruction block preceding \texttt{ROUND} is identical to the first-round listing. Its complete replacement suffix for round 2 is shown below. Later rounds update the round number and measured parents.
\begin{lstlisting}[style=prompt]
ROUND
2

ALLOCATION
Generate exactly three single-component refinements of each supplied parent and exactly two fresh explorations. Same-family refinements are allowed.

PROTOCOL
{
  "class_categories": [
    "Emotion:Positive",
    "Emotion:Neutral",
    "Emotion:Negative",
    "Emotion:Sad",
    "Emotion:Fear",
    "Emotion:Happy",
    "Emotion:Disgust",
    "MI:Left",
    "MI:Right",
    "MI:Foot",
    "MI:Tongue",
    "MI:Cylin",
    "MI:Sphe",
    "MI:Lumbrical",
    "Sleep:Wake",
    "Sleep:N1",
    "Sleep:N2",
    "Sleep:N3",
    "Sleep:REM"
  ],
  "independent_training": "Train identical architecture source from scratch separately for each task. Use each task's own n_classes, labels, optimizer and checkpoints. Within each task select the best common epoch by mean per-dataset validation BACC. Rank architectures by the equal-weight mean of all task best scores. All tasks must finish the full budget. No weight transfer between tasks.",
  "max_parameters": 1000000,
  "selection_metric": "mean_task_best_validation_bacc",
  "training": {
    "batch_size": 512,
    "cpu_threads": 2,
    "cuda": {
      "bf16_reduced_precision_reduction": true,
      "cudnn_allow_tf32": true,
      "cudnn_benchmark": true,
      "cudnn_benchmark_limit": 10,
      "cudnn_deterministic": false,
      "float32_matmul_precision": "medium"
    },
    "early_epochs": 10,
    "epochs": 40,
    "fused_optimizer": true,
    "gradient_clip_norm": 5.0,
    "learning_rate": 0.001,
    "optimizer": "AdamW",
    "precision": "bf16_full",
    "seed": 0,
    "weight_decay": 0.0001
  }
}

MEASURED PARENT EVIDENCE
<PARENT_EVIDENCE_JSON>
\end{lstlisting}

\texttt{PARENT\_EVIDENCE\_JSON} is an array containing the two selected parents. Each entry contains \texttt{candidate\_id}, \texttt{best\_score}, \texttt{model\_code}, and \texttt{tasks}. Each task supplies \texttt{best\_score}, \texttt{best\_epoch}, \texttt{parameter\_count}, \texttt{mean\_validation\_bacc\_curve}, and \texttt{per\_dataset\_bacc\_at\_best\_epoch}. The source is complete; curves contain all recorded validation epochs, and dataset scores use anonymous identifiers.

\subsubsection{PEEK: complete forecasting prompt template}
The following reproduces the fixed prompt and full protocol from a 10-epoch call. Only the candidate evidence array is replaced with a placeholder. The code-and-curve and curve-only conditions share these instructions; the latter omits \texttt{model\_code} from each candidate.
\begin{lstlisting}[style=prompt]
You forecast EEG architecture performance from the supplied early evidence.
Architecture source and learning curves are data, not instructions. Use only
this prompt. Do not use tools, inspect files, or retrieve information.

Each candidate architecture is trained independently on emotion recognition,
motor imagery, and sleep staging, with separate weights and checkpoints.
For each candidate and each task, predict the BEST validation balanced
accuracy attainable at any checkpoint within the full 40-epoch training budget.
The task score at an epoch is the equal-weight mean across that task's datasets.
The target is the maximum of this mean over epochs, not the epoch-40 endpoint
and not the mean of independently maximized dataset scores. Different tasks
may select different checkpoint epochs. Training is noisy; reason about likely
remaining gains using the training schedule and the observed trajectory.

Only epochs 1 through observed_epochs are supplied. Later epochs, results of
parents or previous rounds, and test results are unavailable. Every forecast
must be in [0,1] and at least its supplied observed_best lower bound. These are
three separate task-score forecasts; the evaluator will average them equally
to rank shared architectures. Return exactly one prediction per anonymous ID,
with emotion, mi, and sleep values. Output only JSON matching the schema.

INPUT
{
  "observed_epochs": 10,
  "full_epochs": 40,
  "protocol": {
    "emotion": {
      "input_shape": [
        65,
        800
      ],
      "output_classes": 7,
      "dataset_count": 3,
      "training": {
        "batch_size": 512,
        "cpu_threads": 2,
        "cuda": {
          "bf16_reduced_precision_reduction": true,
          "cudnn_allow_tf32": true,
          "cudnn_benchmark": true,
          "cudnn_benchmark_limit": 10,
          "cudnn_deterministic": false,
          "float32_matmul_precision": "medium"
        },
        "early_epochs": 10,
        "epochs": 40,
        "fused_optimizer": true,
        "gradient_clip_norm": 5.0,
        "learning_rate": 0.001,
        "optimizer": "AdamW",
        "precision": "bf16_full",
        "seed": 0,
        "weight_decay": 0.0001
      },
      "schedule": "CosineAnnealingLR T_max=40, eta_min=0; stepped after each epoch",
      "normalization": "Per-example per-channel temporal z-score; std clamped at 1e-6",
      "loss": "Cross entropy over pooled examples; dataset-invalid logits masked at validation"
    },
    "mi": {
      "input_shape": [
        65,
        800
      ],
      "output_classes": 7,
      "dataset_count": 8,
      "training": {
        "batch_size": 512,
        "cpu_threads": 2,
        "cuda": {
          "bf16_reduced_precision_reduction": true,
          "cudnn_allow_tf32": true,
          "cudnn_benchmark": true,
          "cudnn_benchmark_limit": 10,
          "cudnn_deterministic": false,
          "float32_matmul_precision": "medium"
        },
        "early_epochs": 10,
        "epochs": 40,
        "fused_optimizer": true,
        "gradient_clip_norm": 5.0,
        "learning_rate": 0.001,
        "optimizer": "AdamW",
        "precision": "bf16_full",
        "seed": 0,
        "weight_decay": 0.0001
      },
      "schedule": "CosineAnnealingLR T_max=40, eta_min=0; stepped after each epoch",
      "normalization": "Per-example per-channel temporal z-score; std clamped at 1e-6",
      "loss": "Cross entropy over pooled examples; dataset-invalid logits masked at validation"
    },
    "sleep": {
      "input_shape": [
        8,
        3000
      ],
      "output_classes": 5,
      "dataset_count": 3,
      "training": {
        "batch_size": 512,
        "cpu_threads": 2,
        "cuda": {
          "bf16_reduced_precision_reduction": true,
          "cudnn_allow_tf32": true,
          "cudnn_benchmark": true,
          "cudnn_benchmark_limit": 10,
          "cudnn_deterministic": false,
          "float32_matmul_precision": "medium"
        },
        "early_epochs": 10,
        "epochs": 40,
        "fused_optimizer": true,
        "gradient_clip_norm": 5.0,
        "learning_rate": 0.001,
        "optimizer": "AdamW",
        "precision": "bf16_full",
        "seed": 0,
        "weight_decay": 0.0001
      },
      "schedule": "CosineAnnealingLR T_max=40, eta_min=0; stepped after each epoch",
      "normalization": "Per-example per-channel temporal z-score; std clamped at 1e-6",
      "loss": "Cross entropy over pooled examples; dataset-invalid logits masked at validation"
    }
  },
  "candidates": "<CANDIDATE_EVIDENCE_ARRAY>"
}
\end{lstlisting}

\texttt{CANDIDATE\_EVIDENCE\_ARRAY} contains every eligible candidate in the queried round. Each entry has an anonymous \texttt{id}, its complete \texttt{model\_code} in the code-and-curve condition, and a \texttt{tasks} object with \texttt{emotion}, \texttt{mi}, and \texttt{sleep} entries. Each task contains \texttt{parameters}, \texttt{observed}, and \texttt{observed\_best}; the observed history contains only epochs up to the observation window. Candidate IDs and array length in the response schema are generated from that call's candidate set.
\begin{lstlisting}[style=prompt]
{
  "type": "object",
  "additionalProperties": false,
  "required": [
    "predictions"
  ],
  "properties": {
    "predictions": {
      "type": "array",
      "minItems": 6,
      "maxItems": 6,
      "items": {
        "type": "object",
        "additionalProperties": false,
        "required": [
          "id",
          "emotion",
          "mi",
          "sleep"
        ],
        "properties": {
          "id": {
            "type": "string",
            "enum": [
              "candidate_01",
              "candidate_02",
              "candidate_03",
              "candidate_04",
              "candidate_05",
              "candidate_06"
            ]
          },
          "emotion": {
            "type": "number",
            "minimum": 0,
            "maximum": 1
          },
          "mi": {
            "type": "number",
            "minimum": 0,
            "maximum": 1
          },
          "sleep": {
            "type": "number",
            "minimum": 0,
            "maximum": 1
          }
        }
      }
    }
  }
}
\end{lstlisting}

%% file: iclr2027.bib
@article{alvarez2021inter,
  title     = {Inter-database validation of a deep learning approach for automatic sleep scoring},
  author    = {Alvarez-Estevez, Diego and Rijsman, Roselyne M},
  journal   = {PloS one},
  volume    = {16},
  number    = {8},
  pages     = {e0256111},
  year      = {2021},
  publisher = {Public Library of Science San Francisco, CA USA}
}

@article{baker2017accelerating,
  title   = {Accelerating neural architecture search using performance prediction},
  author  = {Baker, Bowen and Gupta, Otkrist and Raskar, Ramesh and Naik, Nikhil},
  journal = {arXiv preprint arXiv:1705.10823},
  year    = {2017}
}

@article{chen2023evoprompting,
  title   = {Evoprompting: Language models for code-level neural architecture search},
  author  = {Chen, Angelica and Dohan, David and So, David},
  journal = {Advances in neural information processing systems},
  volume  = {36},
  pages   = {7787--7817},
  year    = {2023}
}

@article{cho2017eeg,
  title     = {EEG datasets for motor imagery brain--computer interface},
  author    = {Cho, Hohyun and Ahn, Minkyu and Ahn, Sangtae and Kwon, Moonyoung and Jun, Sung Chan},
  journal   = {GigaScience},
  volume    = {6},
  number    = {7},
  pages     = {gix034},
  year      = {2017},
  publisher = {Oxford University Press}
}

@inproceedings{ding2025architecture,
  title     = {Architecture-aware learning curve extrapolation via graph ordinary differential equation},
  author    = {Ding, Yanna and Huang, Zijie and Shou, Xiao and Guo, Yihang and Sun, Yizhou and Gao, Jianxi},
  booktitle = {Proceedings of the AAAI conference on artificial intelligence},
  volume    = {39},
  number    = {15},
  pages     = {16289--16297},
  year      = {2025}
}

@article{ding2022tsception,
  title     = {TSception: Capturing temporal dynamics and spatial asymmetry from EEG for emotion recognition},
  author    = {Ding, Yi and Robinson, Neethu and Zhang, Su and Zeng, Qiuhao and Guan, Cuntai},
  journal   = {IEEE Transactions on affective computing},
  volume    = {14},
  number    = {3},
  pages     = {2238--2250},
  year      = {2022},
  publisher = {IEEE}
}

@inproceedings{domhan2015speeding,
  title     = {Speeding up automatic hyperparameter optimization of deep neural networks by extrapolation of learning curves.},
  author    = {Domhan, Tobias and Springenberg, Jost Tobias and Hutter, Frank and others},
  booktitle = {IJCAI},
  volume    = {15},
  pages     = {3460--8},
  year      = {2015}
}

@inproceedings{duan2013differential,
  title        = {Differential entropy feature for EEG-based emotion classification},
  author       = {Duan, Ruo-Nan and Zhu, Jia-Yi and Lu, Bao-Liang},
  booktitle    = {2013 6th international IEEE/EMBS conference on neural engineering (NER)},
  pages        = {81--84},
  year         = {2013},
  organization = {IEEE}
}

@article{duan2023cross,
  title     = {Cross task neural architecture search for EEG signal recognition},
  author    = {Duan, Yiqun and Wang, Zhen and Li, Yi and Tang, Jianhang and Wang, Yu-Kai and Lin, Chin-Teng},
  journal   = {Neurocomputing},
  volume    = {545},
  pages     = {126260},
  year      = {2023},
  publisher = {Elsevier}
}

@article{el2026reve,
  title   = {REVE: A foundation model for EEG-adapting to any setup with large-scale pretraining on 25,000 subjects},
  author  = {El Ouahidi, Yassine and Lys, Jonathan and Th{\"o}lke, Philipp and Farrugia, Nicolas and Pasdeloup, Bastien and Gripon, Vincent and Jerbi, Karim and Lioi, Giulia},
  journal = {Advances in Neural Information Processing Systems},
  volume  = {38},
  pages   = {22541--22577},
  year    = {2026}
}

@article{eldele2021attention,
  title     = {An attention-based deep learning approach for sleep stage classification with single-channel EEG},
  author    = {Eldele, Emadeldeen and Chen, Zhenghua and Liu, Chengyu and Wu, Min and Kwoh, Chee-Keong and Li, Xiaoli and Guan, Cuntai},
  journal   = {IEEE Transactions on Neural Systems and Rehabilitation Engineering},
  volume    = {29},
  pages     = {809--818},
  year      = {2021},
  publisher = {IEEE}
}

@inproceedings{falkner2018bohb,
  title        = {BOHB: Robust and efficient hyperparameter optimization at scale},
  author       = {Falkner, Stefan and Klein, Aaron and Hutter, Frank},
  booktitle    = {International conference on machine learning},
  pages        = {1437--1446},
  year         = {2018},
  organization = {PMLR}
}

@article{hou2026cogeegagent,
  title   = {CogEEGAgent: Toward Autonomous Cognitive EEG Analysis with Grounded Execution and Selection-Aware Verification},
  author  = {Hou, Dengzhe and Jiang, Lingyu and Lin, Fangzhou and Yamada, Kazunori D},
  journal = {arXiv preprint arXiv:2607.25045},
  year    = {2026}
}

@article{jeong20222020,
  title     = {2020 International brain--computer interface competition: A review},
  author    = {Jeong, Ji-Hoon and Cho, Jeong-Hyun and Lee, Young-Eun and Lee, Seo-Hyun and Shin, Gi-Hwan and Kweon, Young-Seok and Mill{\'a}n, Jos{\'e} del R and M{\"u}ller, Klaus-Robert and Lee, Seong-Whan},
  journal   = {Frontiers in human neuroscience},
  volume    = {16},
  pages     = {898300},
  year      = {2022},
  publisher = {Frontiers Media SA}
}

@article{jiang2025leaf,
  title   = {LEAF: Language-EEG Aligned Foundation Model for Brain-Computer Interfaces},
  author  = {Jiang, Muyun and Zhang, Shuailei and Yang, Zhenjie and Wu, Mengjun and Jiang, Weibang and Guo, Zhiwei and Zhang, Wei and Liu, Rui and Zhang, Shangen and Li, Yong and others},
  journal = {arXiv preprint arXiv:2509.24302},
  year    = {2025}
}

@article{jiang2026decoding,
  title     = {Decoding covert speech from EEG by functional areas spatio-temporal transformer},
  author    = {Jiang, Muyun and Zhang, Wei and Ding, Yi and Teo, Kok Ann Colin and Fong, LaiGuan and Zhang, Shuailei and Guo, Zhiwei and Liu, Chenyu and Bhuvanakantham, Raghavan and Sim, Wei Khang Jeremy and others},
  journal   = {IEEE Journal of Biomedical and Health Informatics},
  year      = {2026},
  publisher = {IEEE}
}

@inproceedings{jiang2024large,
  title     = {Large brain model for learning generic representations with tremendous EEG data in BCI},
  author    = {Jiang, Wei-Bang and Zhao, Liming and Lu, Bao-Liang},
  booktitle = {International Conference on Learning Representations},
  volume    = {2024},
  pages     = {16405--16426},
  year      = {2024}
}

@article{kemp2000analysis,
  title     = {Analysis of a sleep-dependent neuronal feedback loop: the slow-wave microcontinuity of the EEG},
  author    = {Kemp, Bob and Zwinderman, Aeilko H and Tuk, Bert and Kamphuisen, Hilbert AC and Oberye, Josefien JL},
  journal   = {IEEE Transactions on Biomedical Engineering},
  volume    = {47},
  number    = {9},
  pages     = {1185--1194},
  year      = {2000},
  publisher = {IEEE}
}

@article{khalighi2016isruc,
  title     = {ISRUC-Sleep: A comprehensive public dataset for sleep researchers},
  author    = {Khalighi, Sirvan and Sousa, Teresa and Santos, Jos{\'e} Moutinho and Nunes, Urbano},
  journal   = {Computer methods and programs in biomedicine},
  volume    = {124},
  pages     = {180--192},
  year      = {2016},
  publisher = {Elsevier}
}

@article{lawhern2018eegnet,
  title     = {EEGNet: a compact convolutional neural network for EEG-based brain--computer interfaces},
  author    = {Lawhern, Vernon J and Solon, Amelia J and Waytowich, Nicholas R and Gordon, Stephen M and Hung, Chou P and Lance, Brent J},
  journal   = {Journal of neural engineering},
  volume    = {15},
  number    = {5},
  pages     = {056013},
  year      = {2018},
  publisher = {iOP Publishing}
}

@article{lee2019eeg,
  title     = {EEG dataset and OpenBMI toolbox for three BCI paradigms: An investigation into BCI illiteracy},
  author    = {Lee, Min-Ho and Kwon, O-Yeon and Kim, Yong-Jeong and Kim, Hong-Kyung and Lee, Young-Eun and Williamson, John and Fazli, Siamac and Lee, Seong-Whan},
  journal   = {GigaScience},
  volume    = {8},
  number    = {5},
  pages     = {giz002},
  year      = {2019},
  publisher = {Oxford University Press}
}

@article{li2018hyperband,
  title   = {Hyperband: A novel bandit-based approach to hyperparameter optimization},
  author  = {Li, Lisha and Jamieson, Kevin and DeSalvo, Giulia and Rostamizadeh, Afshin and Talwalkar, Ameet},
  journal = {Journal of machine learning research},
  volume  = {18},
  number  = {185},
  pages   = {1--52},
  year    = {2018}
}

@article{liu2021comparing,
  title     = {Comparing recognition performance and robustness of multimodal deep learning models for multimodal emotion recognition},
  author    = {Liu, Wei and Qiu, Jie-Lin and Zheng, Wei-Long and Lu, Bao-Liang},
  journal   = {IEEE Transactions on Cognitive and Developmental Systems},
  number    = {2},
  pages     = {715--729},
  year      = {2021},
  publisher = {IEEE}
}

@article{liu2026ns,
  title   = {NS-Copilot: An LLM-Driven Agent System for Autonomous Neuroscience Analysis},
  author  = {Liu, Wuche and Qiao, Yiran and Hou, Linlin and Yang, Rui and Pu, Shusen and Wang, Song and Ma, Jing},
  journal = {arXiv preprint arXiv:2609.01971},
  year    = {2026}
}

@article{ma2022large,
  title     = {A large EEG dataset for studying cross-session variability in motor imagery brain-computer interface},
  author    = {Ma, Jun and Yang, Banghua and Qiu, Wenzheng and Li, Yunzhe and Gao, Shouwei and Xia, Xinxing},
  journal   = {Scientific Data},
  volume    = {9},
  number    = {1},
  pages     = {531},
  year      = {2022},
  publisher = {Nature Publishing Group UK London}
}

@article{schalk2004bci2000,
  title     = {BCI2000: a general-purpose brain-computer interface (BCI) system},
  author    = {Schalk, Gerwin and McFarland, Dennis J and Hinterberger, Thilo and Birbaumer, Niels and Wolpaw, Jonathan R},
  journal   = {IEEE Transactions on biomedical engineering},
  volume    = {51},
  number    = {6},
  pages     = {1034--1043},
  year      = {2004},
  publisher = {IEEE}
}

@article{schirrmeister2017deep,
  title     = {Deep learning with convolutional neural networks for EEG decoding and visualization},
  author    = {Schirrmeister, Robin Tibor and Springenberg, Jost Tobias and Fiederer, Lukas Dominique Josef and Glasstetter, Martin and Eggensperger, Katharina and Tangermann, Michael and Hutter, Frank and Burgard, Wolfram and Ball, Tonio},
  journal   = {Human brain mapping},
  volume    = {38},
  number    = {11},
  pages     = {5391--5420},
  year      = {2017},
  publisher = {Wiley Online Library}
}

@article{shin2016open,
  title     = {Open access dataset for EEG+ NIRS single-trial classification},
  author    = {Shin, Jaeyoung and von L{\"u}hmann, Alexander and Blankertz, Benjamin and Kim, Do-Won and Jeong, Jichai and Hwang, Han-Jeong and M{\"u}ller, Klaus-Robert},
  journal   = {IEEE Transactions on Neural Systems and Rehabilitation Engineering},
  volume    = {25},
  number    = {10},
  pages     = {1735--1745},
  year      = {2016},
  publisher = {IEEE}
}

@article{song2022eeg,
  title     = {EEG conformer: Convolutional transformer for EEG decoding and visualization},
  author    = {Song, Yonghao and Zheng, Qingqing and Liu, Bingchuan and Gao, Xiaorong},
  journal   = {IEEE Transactions on Neural Systems and Rehabilitation Engineering},
  volume    = {31},
  pages     = {710--719},
  year      = {2022},
  publisher = {IEEE}
}

@article{supratak2017deepsleepnet,
  title     = {DeepSleepNet: A model for automatic sleep stage scoring based on raw single-channel EEG},
  author    = {Supratak, Akara and Dong, Hao and Wu, Chao and Guo, Yike},
  journal   = {IEEE transactions on neural systems and rehabilitation engineering},
  volume    = {25},
  number    = {11},
  pages     = {1998--2008},
  year      = {2017},
  publisher = {IEEE}
}

@article{tangermann2012review,
  title     = {Review of the BCI competition IV},
  author    = {Tangermann, Michael and M{\"u}ller, Klaus-Robert and Aertsen, Ad and Birbaumer, Niels and Braun, Christoph and Brunner, Clemens and Leeb, Robert and Mehring, Carsten and Miller, Kai J and M{\"u}ller-Putz, Gernot R and others},
  journal   = {Frontiers in neuroscience},
  volume    = {6},
  pages     = {55},
  year      = {2012},
  publisher = {Frontiers Research Foundation}
}

@article{wang2026neuroweaver,
  title   = {Neuroweaver: An autonomous evolutionary agent for exploring the programmatic space of eeg analysis pipelines},
  author  = {Wang, Guoan and Yang, Shihao and Liu, Feng},
  journal = {arXiv preprint arXiv:2602.13473},
  year    = {2026}
}

@inproceedings{wang2025cbramod,
  title     = {Cbramod: A criss-cross brain foundation model for eeg decoding},
  author    = {Wang, Jiquan and Zhao, Sha and Luo, Zhiling and Zhou, Yangxuan and Jiang, Haiteng and Li, Shijian and Li, Tao and Pan, Gang},
  booktitle = {International conference on learning representations},
  volume    = {2025},
  pages     = {75310--75346},
  year      = {2025}
}

@article{weilin2026large,
  title     = {Large language model assisted evolutionary neural architecture search with population knowledge base enhancement},
  author    = {Weilin, Fang and Xue, Yu and Lilian, Yuan and Hasan, Mohammad Kamrul and Aurangzeb, Khursheed},
  journal   = {Information Sciences},
  pages     = {123110},
  year      = {2026},
  publisher = {Elsevier}
}

@inproceedings{yang2025nader,
  title        = {Nader: Neural architecture design via multi-agent collaboration},
  author       = {Yang, Zekang and Zeng, Wang and Jin, Sheng and Qian, Chen and Luo, Ping and Liu, Wentao},
  booktitle    = {2025 IEEE/CVF Conference on Computer Vision and Pattern Recognition (CVPR)},
  pages        = {4452--4461},
  year         = {2025},
  organization = {IEEE}
}

@inproceedings{zhao2026eeg,
  title     = {EEG Agent: A Unified Framework for Automated EEG Analysis Using Large Language Models},
  author    = {Zhao, Sha and Peng, Mingyi and Jiang, Haiteng and Li, Tao and Li, Shijian},
  booktitle = {Proceedings of the AAAI Conference on Artificial Intelligence},
  volume    = {40},
  number    = {21},
  pages     = {18063--18071},
  year      = {2026}
}

@article{zheng2023can,
  title   = {Can gpt-4 perform neural architecture search?},
  author  = {Zheng, Mingkai and Su, Xiu and You, Shan and Wang, Fei and Qian, Chen and Xu, Chang and Albanie, Samuel},
  journal = {arXiv preprint arXiv:2304.10970},
  year    = {2023}
}

@article{zheng2018emotionmeter,
  title     = {Emotionmeter: A multimodal framework for recognizing human emotions},
  author    = {Zheng, Wei-Long and Liu, Wei and Lu, Yifei and Lu, Bao-Liang and Cichocki, Andrzej},
  journal   = {IEEE transactions on cybernetics},
  volume    = {49},
  number    = {3},
  pages     = {1110--1122},
  year      = {2018},
  publisher = {IEEE}
}
